\documentclass{article}
\usepackage{amsmath,amssymb}
\usepackage{graphicx,subfigure}
\usepackage{algorithm,algorithmic}

\usepackage[left=1in,right=1in]{geometry}
\graphicspath{{../}}

\newcommand{\RR}{\mathbf{R}}
\newcommand{\T}{^\top}
\newcommand{\calT}{\mathcal{T}}

\newcommand{\vx}{\boldsymbol{x}}

\newcommand{\vchi}{\boldsymbol{\chi}}
\newcommand{\vtheta}{\boldsymbol{\theta}}
\newcommand{\va}{\boldsymbol{a}}

\newcommand{\argmin}{\mathop{\rm argmin}}
\newcommand{\argmax}{\mathop{\rm argmax}}

\begin{document}

\title{Discovering Emerging Topics in Social Streams via Link Anomaly
Detection}

\author{
Toshimitsu Takahashi\\
Institute of Industrial Science\\
The University of Tokyo\\
Tokyo, Japan\\
Email: takahashi@tauhat.com
\and
Ryota Tomioka\\
Department of Mathematical Informatics\\
The University of Tokyo\\
Tokyo, Japan\\
Email: tomioka@mist.i.u-tokyo.ac.jp
\and
Kenji Yamanishi\\
Department of Mathematical Informatics\\
The University of Tokyo\\
Tokyo, Japan\\
Email: yamanishi@mist.i.u-tokyo.ac.jp
}

\maketitle
\begin{abstract}
Detection of emerging topics are now receiving renewed interest
motivated by the rapid growth of social networks. Conventional
term-frequency-based approaches may not be appropriate in this context,
because the information exchanged are not only texts but also images,
URLs, and videos. We focus on the social aspects of theses
networks. That is, the links between users that are generated
dynamically intentionally or unintentionally through replies, mentions,
and retweets. We propose a probability model of the mentioning
behaviour of a social network user, and propose to detect the emergence
of a new topic from the anomaly measured through the model. 
We combine the proposed mention anomaly score
 with a recently proposed
change-point detection technique based on the Sequentially
Discounting Normalized Maximum Likelihood (SDNML), or with Kleinberg's
 burst model.
Aggregating anomaly scores from hundreds of users, we
show that we can detect emerging topics only based on the
reply/mention relationships in social network posts. We demonstrate our
technique in a number of real data sets we gathered from Twitter. The
 experiments show that the proposed mention-anomaly-based approaches can 
detect new topics at least as early as the conventional term-frequency-based
approach, and sometimes much earlier when the keyword is ill-defined.
\end{abstract}


\section{Introduction}
Communication through social networks, such as Facebook and Twitter, is increasing its importance in
our daily life. Since the information exchanged over social networks are
not only texts but also URLs, images, and videos, they are challenging
test beds for the study of data mining.

There is another type of information that is intentionally or
unintentionally exchanged over social networks: mentions. Here we mean by
mentions {\em links} to other users of the same social network in the form of
message-to, reply-to, retweet-of, or explicitly in the text. One post
may contain a number of mentions. Some users may include mentions in their posts
rarely; other users may be mentioning their friends all the time.  Some
users (like celebrities) may receive mentions every minute; for others,
being mentioned might be a rare occasion. In this sense, {\em mention is like
a language} with the number of words equal to the number of users in
a social network.

We are interested in detecting emerging topics from social
network streams based on monitoring the mentioning behaviour of users.
Our basic assumption is that a new (emerging) topic is something people
feel like discussing about, commenting about, or forwarding the
information further to their friends.
Conventional approaches for topic detection have mainly been concerned
with the frequencies of (textual) words~\cite{AllCarDodYamYanoth98,Kle03}. A term frequency based
approach could suffer from the ambiguity caused by synonyms or
homonyms. It may also require complicated preprocessing (e.g., segmentation)
depending on the target language. Moreover, it cannot be applied when
the contents of the messages are mostly non-textual information.
On the other hands, the ``words'' formed by mentions are unique,
requires little prepossessing to obtain (the information is often
separated from the contents), and are available regardless of the nature
of the contents. 

\begin{figure}[tb]
\begin{center}
 \includegraphics[width=.5\textwidth]{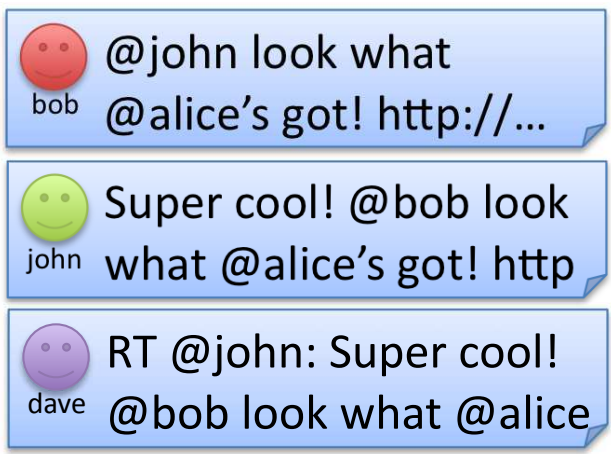}
 \caption{Example of the emergence of a topic in social streams.}
 \label{fig:conversation}
\end{center} 
\end{figure}

Figure~\ref{fig:conversation} shows an example of the emergence of a
topic through posts on social networks. The first post by Bob contains
mentions to Alice and John, which are both probably friends of Bob's; so
there is nothing unusual here. The second post by John is a reply to Bob
but it is also visible to many friends of John's that are not direct
friends of Bob's. Then in the third post, Dave, one of John's friends,
forwards (called retweet in Twitter) the information further down to
his own friends. It is worth mentioning that it is not clear
what the topic of this conversation is about from the textual information,
because they are talking about something (a new gadget, car, or jewelry)
that is shown as a link in the text.

In this paper, we propose a probability model that can capture the normal
mentioning behaviour of a user, which consists of both the
number of mentions per post and the frequency of users occurring in the mentions.
Then this model is used to measure the {\em anomaly} of future user
behaviour. Using the proposed probability model, we can quantitatively
measure the novelty or possible impact of a post reflected in the
mentioning behaviour of the user.
We aggregate the anomaly scores obtained in this way over
hundreds of users  and apply a recently proposed change-point detection
technique based on the Sequentially Discounting Normalized Maximum
Likelihood (SDNML) coding~\cite{UraYamTomIwa11}. This technique can
detect a change in the statistical dependence structure in the time
series of aggregated anomaly scores, and pin-point where the topic
emergence is; see Figure~\ref{fig:system}.  The effectiveness of
the proposed approach is demonstrated on four data sets we have collected
from Twitter. We show that our approach can detect the emergence of a
new topic at least as fast as using the best term that was not obvious
at the moment. Furthermore, we show that in two out of four data sets, the proposed link-anomaly based method can detect the emergence
of the topics earlier than keyword-frequency based methods, which can be
explained by the keyword ambiguity we mentioned above.

\section{Related work}
Detection and tracking of topics have been studied extensively in the
area of topic detection and tracking
(TDT)~\cite{AllCarDodYamYanoth98}. In this context, the main task is to
either classify a new document into one of the known topics (tracking)
or to detect that it belongs to none of the known
categories. Subsequently, temporal structure of topics have been modeled
and analyzed through dynamic model selection~\cite{MorYam04}, temporal
text mining~\cite{MeiZha05}, and factorial hidden Markov
models~\cite{KraLesGue06}.

Another line of research is concerned with formalizing the notion of
``bursts'' in a stream of documents. In his seminal paper, Kleinberg
modeled bursts using time varying Poisson process with a hidden discrete
process that controls the firing rate~\cite{Kle03}. Recently, He and
Parker developed a physics inspired model of bursts based
on the change in the momentum of topics~\cite{HePar10}.

All the above mentioned studies make use of textual content of the
documents, but not the social content of the documents. The social
content (links) have been utilized in the study of citation
networks~\cite{Sma99}. However, citation networks are often analyzed in
a stationary setting.

The novelty of the current paper lies in focusing on the social content
of the documents (posts) and in combining this with a change-point analysis.

\section{Proposed Method}
\begin{figure}[tb]
 \begin{center}
  \includegraphics[width=.7\textwidth]{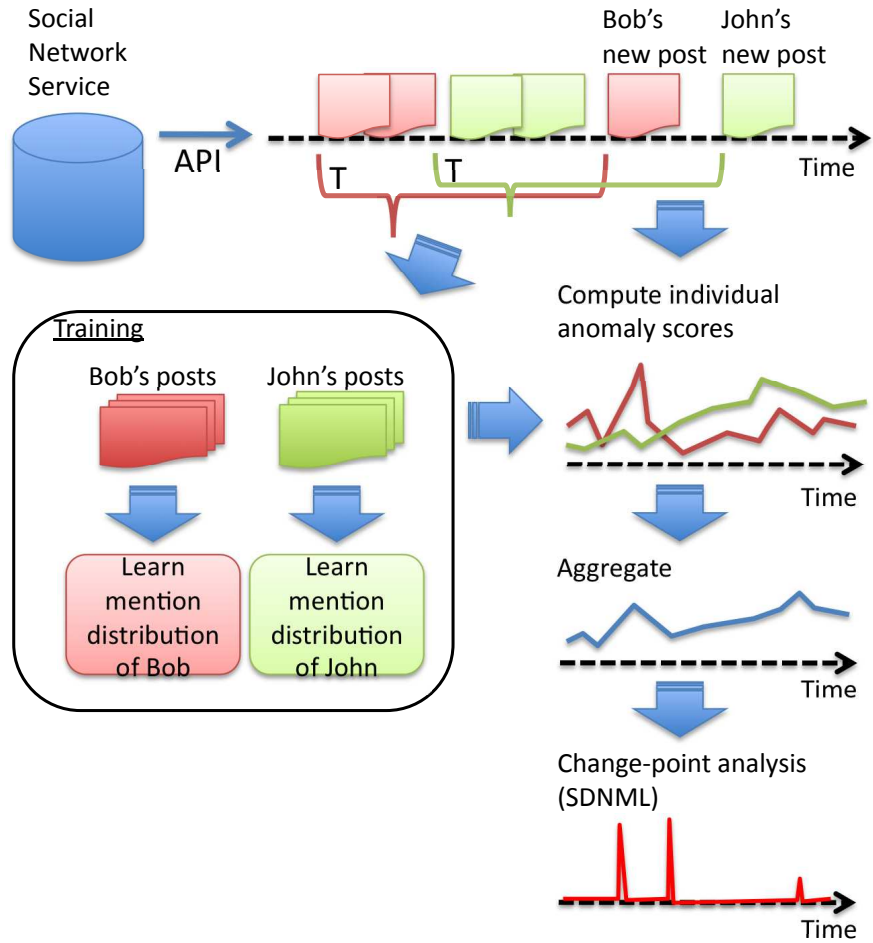}
  \caption{Overall flow of the proposed method.}
  \label{fig:system}
 \end{center}
\end{figure}

The overall flow of the proposed method is shown in Figure~\ref{fig:system}.
We assume that the data arrives from a social network service in a
sequential manner through some API. For each new post we use samples
within the past $T$ time interval for the corresponding user for
training the mention model we propose below. We assign anomaly score to
each post based on the learned probability distribution. The score is
then aggregated over users and further fed into a change-point analysis.

\subsection{Probability Model}
We characterize a post in a social network stream by the number of mentions $k$ it contains, and
the set $V$ of names (IDs) of the users mentioned in the
post. Formally, we consider the following joint probability distribution
\begin{align}
\label{eq:model}
 P(k,V|\theta,\{\pi_v\} )&=P(k|\theta)\prod_{v\in V}\pi_v.
\end{align}
Here the joint distribution consists of two parts: the probability of
the number of mentions $k$ and the probability of each mention given the
number of mentions. The probability of the number of mentions
$P(k|\theta)$ is defined as a geometric distribution with parameter
$\theta$ as follows: 
\begin{align}
\label{eq:geom}
P(k|\theta)&=(1-\theta)^k\theta.
\end{align}
On the other hand, the probability of mentioning users in $V$ is defined
as independent, identical multinomial distribution with parameters
$\pi_v$ ($\sum_{v}\pi_v=1$).

Suppose that we are given $n$ training examples
$\calT=\{(k_1,V_1)$, $\ldots$, $(k_n,V_n)\}$ from which we would like to learn the
predictive distribution 
\begin{align}
\label{eq:pred}
P(k,V|\calT) &= P(k|\calT)\prod_{v\in V}P(v|\calT).
\end{align}
First we compute the predictive distribution with respect to the the
number of mentions $P(k|\calT)$. This can be obtained by
assuming a beta distribution as a prior and integrating out the
parameter $\theta$. The density function of the beta prior distribution
is written as follows:
\begin{align*}
p(\theta|\alpha,\beta)=\frac{(1-\theta)^{\beta-1}\theta^{\alpha-1}}{B(\alpha,\beta)},
\end{align*}
where $\alpha$ and $\beta$ are parameters of the beta distribution and
$B(\alpha,\beta)$ is the beta function. By the Bayes rule, the predictive
distribution can be obtained as follows:
\begin{align*}
 P(k|\calT,\alpha,\beta)&=
P(k|k_1,\ldots,k_n,\alpha,\beta)\\
&=\frac{P(k,k_1,\ldots,k_n|\alpha,\beta)}{P(k_1,\ldots,k_n|\alpha,\beta)}\\
&=\frac{\int_{0}^{1}(1-\theta)^{\sum_{i=1}^nk_i+k+\beta-1}\theta^{n+1+\alpha-1}{\rm  d}\theta}{\int_{0}^{1}(1-\theta)^{\sum_{i=1}^nk_i+\beta-1}\theta^{n+\alpha-1}{\rm  d}\theta }.
\end{align*}
Both the integrals on the numerator and denominator can be obtained in
closed forms as beta functions and the predictive distribution can be rewritten as follows:
\begin{align*}
 P(k|\calT,\alpha,\beta)&=
\frac{B(n+1+\alpha,\sum_{i=1}^nk_i+k+\beta)}{B(n+\alpha,\sum_{i=1}^nk_i+\beta)}.
\end{align*}
Using the relation between beta function and gamma function, we can
further simplify the expression as follows:
\begin{align}
\label{eq:pred-k}
P(k|\calT,\alpha,\beta)&=
\frac{n+\alpha}{m+k+\beta}\prod_{j=0}^{k}
\frac{m+\beta+j}{n+m+\alpha+\beta+j},
\end{align}
where $m=\sum_{i=1}^{n}k_i$ is the total number of
mentions in the training set $\calT$.

Next, we derive the predictive distribution $P(v|\calT)$ of mentioning
user $v$. The maximum likelihood (ML) estimator is given as
$P(v|\calT)=m_{v}/m$, where $m$ is the number of total
mentions and $m_{v}$ is the number of mentions to user $v$ in the data
set $\calT$. The ML estimator, however, cannot handle users that did not
appear in the training set $\calT$; it would assign probability zero to
all these users, which would appear infinitely anomalous in our
framework. Instead we use the Chinese Restaurant Process (CRP; see \cite{Ald85}) based
estimation. The CRP based estimator assigns probability to each user $v$
that is proportional to the number of mentions $m_v$ in the training
set $\calT$; in addition, it keeps probability proportional to $\gamma$
for mentioning someone who was not mentioned in the training set
$\calT$. Accordingly the probability of known users is given as follows:
\begin{align}
\label{eq:pred-v1}
 P(v|\calT)&=\frac{m_{v}}{m+\gamma} \quad \textrm{(for $v$: $m_v\geq 1$)}.
\end{align}
On the other hand, the probability of mentioning a new user is given as follows:
\begin{align}
\label{eq:pred-v2}
\hspace*{-8mm} P(\{v:m_v=0\}|\calT)&=\frac{\gamma}{m+\gamma}.
\end{align}
\subsection{Computing the link-anomaly score}
In order to compute the anomaly score of a new post $\vx=(t,u,k,V)$ by user
$u$ at time $t$ containing $k$ mentions to users $V$, we compute the probability \eqref{eq:pred} with the
training set $\calT_u^{(t)}$, which is the collection of posts by user
$u$ in the time period $[t-T,t]$ (we use $T=30$ days in this paper). Accordingly the
link-anomaly score is defined as follows:
\begin{align}
s(\vx)&=-\log\left(P(k|\calT_u^{(t)})\prod_{v\in
 V}P(v|\calT_u^{(t)})\right)\nonumber\\
 \label{eq:link-score}
 &=-\log P(k|\calT_u^{(t)})-\sum_{v\in V}\log P(v|\calT_u^{(t)}).
\end{align}
The two terms in the above equation can be computed via
the predictive distribution of the number of mentions \eqref{eq:pred-k},
and the predictive distribution of the
mentionee~\eqref{eq:pred-v1}--\eqref{eq:pred-v2}, respectively.

\subsection{Combining Anomaly Scores from Different Users}
The anomaly score in \eqref{eq:link-score} is computed for each user
depending on the current post of user $u$ and his/her past behaviour
$\calT_u^{(t)}$. In order to measure the general trend of user
behaviour, we propose to 
aggregate the anomaly scores obtained for posts
$\vx_1,\ldots,\vx_n$ using a discretization of window size $\tau>0$ as follows:
\begin{align}
\label{eq:agg-s}
 s_j'=\frac{1}{\tau}\sum_{t_i\in [\tau(j-1),\tau j]}s(\vx_i),
\end{align}
where $\vx_i=(t_i,u_i,k_i,V_i)$ is the post at time $t_i$ by user $u_i$
including $k_i$ mentions to users $V_i$.

\subsection{Change-point detection via Sequentially Discounting Normalized Maximum Likelihood Coding}
\label{sec:change-point}
Given an  aggregated measure of anomaly~\eqref{eq:agg-s}, we apply a
change-point detection technique based on the SDNML
coding~\cite{UraYamTomIwa11}. This 
technique detects a change in the statistical dependence structure of a time series by monitoring the compressibility
of the new piece of data. The sequential version of normalized maximum
likelihood (NML) coding is employed as a coding criterion. More precisely, a
change point is detected through two layers of scoring processes (see also
\cite{YamTak02,TakYam06}); in each layer, the SDNML code length based on
an autoregressive (AR) model is used as a criterion for scoring. Although
the NML code length is known to be optimal~\cite{Ris02}, it is often hard to
compute. The SNML proposed in \cite{RisRooMyl10} is an approximation to
the NML code length that can be computed in a sequential manner. The
SDNML proposed in \cite{UraYamTomIwa11} further employs discounting in
the learning of the AR models.

Algorithmically, the change point detection procedure can be outlined as
follows. For convenience, we denote the aggregate anomaly score as $x_j$
instead of $s_j'$.
\begin{description}
 \item[1. 1st layer learning]
 Let $x^{j-1}:=\{x_1,\ldots,x_{j-1}\}$ be
	    the collection of aggregate anomaly scores from discrete
	    time $1$ to $j-1$. Sequentially learn the SDNML density
	    function $p_{\mbox{\tiny SDNML}}(x_j|x^{j-1})$  ($j=1,2,\ldots$); see
	    Appendix~\ref{sec:sdnml} for details.
 \item[2. 1st layer scoring]
 Compute the intermediate
	    change-point score by smoothing the log loss of the SDNML
	    density function with window size $\kappa$ as follows:
\begin{align*}
y_j &=\frac{1}{\kappa}\sum_{j'=j-\kappa+1}^{j}\left(-\log p_{\mbox{\tiny
 SDNML}}(x_j|x^{j-1})\right).
\end{align*}
\item[4. 2nd layer learning]
 Let $y^{j-1}:=\{y_1,\ldots,y_{j-1}\}$ be
	   the collection of smoothed change-point score obtained as
	   above. Sequentially learn the second layer SDNML density
	   function $p_{\mbox{\tiny SDNML}}(y_j|y^{j-1})$ ($j=1,2,\ldots$); see
	   Appendix~\ref{sec:sdnml} for details.
\item[5. 2nd layer scoring]
 Compute the final change-point score by
	   smoothing the log loss of the SDNML density function as follows:
\begin{align}
\label{eq:score}
 Score(y_j)&=\frac{1}{\kappa}\sum_{j'=j-\kappa+1}^{j}\left(-\log
 p_{\mbox{\tiny SDNML}}(y_j|y^{j-1})\right).
\end{align}
\end{description}

\subsection{Dynamic Threshold Optimization (DTO)}
\label{sec:dto}
We make an alarm if the change-point score exceeds a threshold,
which was determined adaptively using the method of dynamic threshold
optimization~(DTO), proposed in \cite{Yamanishi2005}.

In DTO, we use a 1-dimensional histogram for the representation
of the score distribution. We learn it in a sequential and
discounting way.
Then, for a specified value $\rho$, to determine the threshold to be the
largest score value such that the tail probability beyond the
value does not exceed $\rho$.
We call $\rho$ a {\em threshold parameter}.

The details of DTO are summarized as follows:
Let $N_H$ be a given positive
integer. Let $\{ q(h)(h = 1,\dots ,N_{H}) : 
\sum _{h=1}^{N_{H}} q(h) = 1\}$ be a 1-
dimensional histogram with $N_{H}$ bins where $h$ is an index of
bins, with a smaller index indicating a bin having a smaller
score. For given $a, b$ such that $a < b$, $N_{H}$ bins in the histogram
are set as: $\{(-\infty , a); [a + \{(b - a)/(N_{H}-2)\}\ell , 
[a + \{(b - a)/(N_{H}-2)\}(\ell +1)(\ell = 0,1,...,N_{H}- 3)$ and $[b, \infty )$.
Let $\{q^{(j)}(h)\}$ be a histogram updated after seeing the $j$th score.
The procedures of updating the histogram and DTO are given in Algorithm \ref{fig:DTO}.
\begin{algorithm}[tb]
\caption{Dynamic Threshold Optimization (DTO) \cite{Yamanishi2005}}
  \label{fig:DTO}
{\small 
\begin{algorithmic}
\STATE {\bf \underline{Given:}}  $\{Score_{j}|j=1,2,\ldots\}$: scores,   $N_{H}$: total number of cells,
   $\rho$: parameter for threshold,
   $\lambda_{H}$: estimation parameter,
   $r_{H}$: discounting parameter,
   $M$: data size\\
\STATE {\bf \underline{Initialization:}} Let $q_{1}^{(1)}(h)$ (a weighted
 sufficient statistics) be a uniform distribution.
\FOR{$j=1,\ldots,M-1$}
\STATE {\bf \underline{Threshold optimization:}}
 Let $l$ be the least index such that
   $\sum_{h=1}^{l}q^{(j)}(h)\geq 1- \rho$.
   The threshold at time $j$ is given as 
$$\eta(j) = a + \frac{b-a}{N_{H}-2}(l+1).$$
\STATE {\bf\underline{Alarm output:}} Raise an alarm if 
$Score_{j}\geq \eta(j)$.
\STATE {\bf \underline{Histogram update:}}
\begin{align*}
q_{1}^{(j+1)}(h) &=
 \begin{cases}
(1-r_{H})q_{1}^{(j)}(h)+r_{H}  &
\begin{array}{p{2cm}}
if $Score_{j}$ falls into the $h$th
  cell,
\end{array}\\
(1-r_{H})q_{1}^{(j)}(h) & \text{otherwise.}
 \end{cases}
\end{align*}
\STATE    $q^{(j+1)}(h)\!\!=\!\!(q_{1}^{(j+1)}(h)+\lambda_{H})/(\sum_{h}q_{1}^{(j+1)}(h)+N_{H}\lambda_{H})$.
\ENDFOR
\end{algorithmic} 
}
\end{algorithm}

\section{Experiments}
\subsection{Experimental setup}
We collected four data sets from Twitter. Each data set is
associated with a list of posts in a service called
Togetter\footnote{http://togetter.com/}; Togetter is a
collaborative service where people can tag Twitter posts that are
related to each other and organize a list of posts that belong to a
certain topic. Our goal is to evaluate whether the proposed approach can
detect the emergence of the topics recognized and collected by
people. We have selected four data sets, ``Job hunting'',
``Youtube'', ``NASA'', ``BBC'' each corresponding to a user organized list in
Togetter. For each data set we collected posts from users that appeared in
each list (participants). The number of participants in each data set is
different; see Table~\ref{tab:dataset}.

We compared our proposed approach with a keyword-based change-point detection
method. In the keyword-based method, we looked at a sequence
of occurrence frequencies (observed within one minute) of a keyword
related to the topic; the keyword was manually selected to best capture
the topic. Then we applied DTO described in Section~\ref{sec:dto} 
to the sequence of keyword frequency. In our experience, the sparsity of
the keyword frequency seems to be a bad combination with the SDNML
method; therefore we did not use SDNML in the keyword-based method.
We use the smoothing parameter $\kappa=15$, and the
order of the AR model 30 in the experiments; the parameters in DTO was set as 
$\rho=0.05$, $N_{H}=20$, $\lambda_{H}=0.01$, $r_{H}=0.005$.

Furthermore, we have implemented a two-state version of Kleinberg's
burst detection model~\cite{Kle03} using link-anomaly
score~\eqref{eq:agg-s} and keyword frequency (as in the keyword-based
change-point analysis) to filter out relevant posts. For the
link-anomaly score, we used a threshold to filter out posts to include in
the burst analysis. For the keyword frequency, we used all posts that
include the keyword for the burst analysis. We used the firing rate
parameter of the Poisson point process $0.001$ (1/s) for the non-burst
state and $0.01$ (1/s) for the burst state, and the transition probability
$p=0.3$. We consider the transition from the non-burst state to the
burst state as an ``alarm''.

A drawback of the keyword-based methods (dynamic thresholding and
burst detection) is that the keyword related to the topic must be known in advance, although this is not always the case in practice.
The change-point detected by the keyword-based methods can be thought of as the time when the topic really emerges.
Hence our goal is to detect emerging topics as early as the keyword based methods.


\begin{table}[t]
 \begin{center}
 \caption{Number of participants in each data set.}
 \label{tab:dataset}
  \begin{tabular}[t]{|c|c|}
   \hline
   data set   &$\sharp$ of
participants\\\hline\hline
   ``Job hunting''&200\\\hline
   ``Youtube''&160\\\hline
   ``NASA''&90\\\hline
   ``BBC'' & 47 \\\hline
  \end{tabular}
 \end{center}
\end{table}

\subsection{``Job hunting'' data set}
This data set is related to a controversial post by a famous person in
Japan that ``the reason students having difficulty finding jobs is,
because they are stupid'' and various replies to that post.

\begin{figure}[tb]
  \begin{center}
 \subfigure[Link-anomaly-based change-point analysis. Green: Aggregated anomaly
  score~\eqref{eq:agg-s} at $\tau=1$ minute. Blue: Change-point
  score~\eqref{eq:score}. Red: Alarm time.]{
   \includegraphics[width=.5\columnwidth]{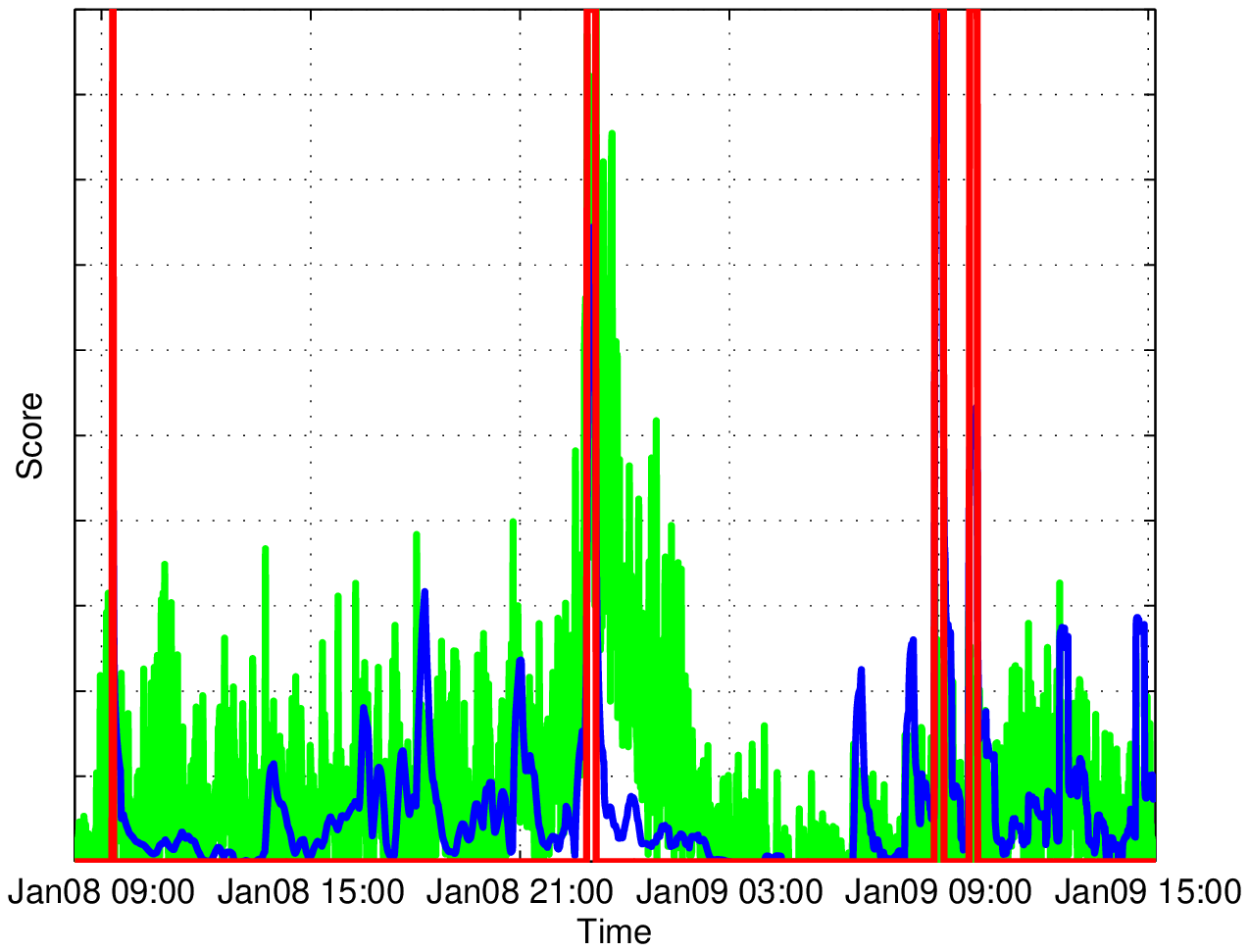}
\label{fig:ikeda}}~\subfigure[Link-anomaly-based burst
   detection.  Blue: Aggregated anomaly
  score~\eqref{eq:agg-s} at $\tau=1$ second.
 Cyan: threshold for the filtering step in Kleinberg's burst model.
 Red: Burst state.]{\includegraphics[width=.5\columnwidth]{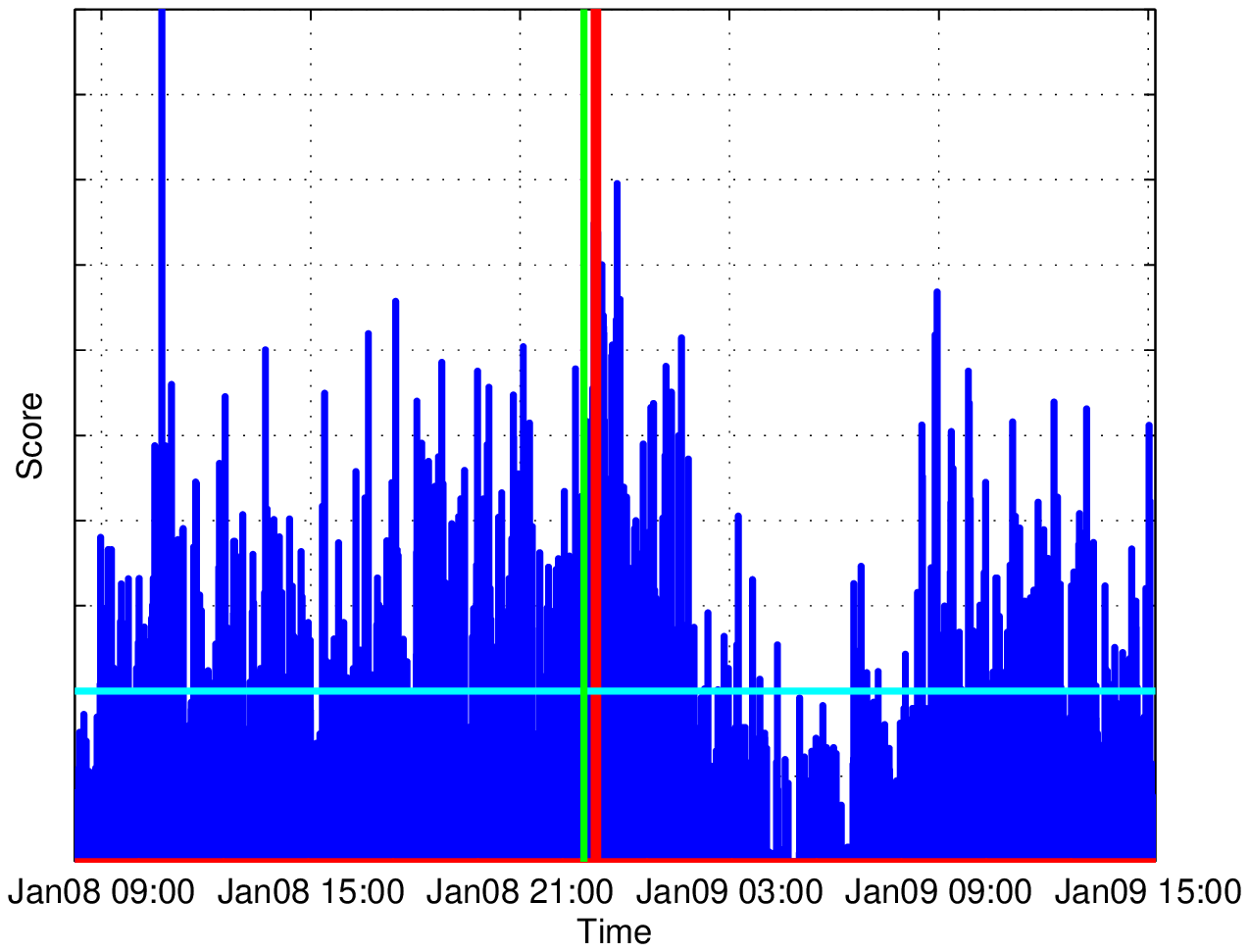}\label{fig:ikeda_link_burst}}
   \subfigure[Keyword-frequency-based change-point analysis.  Blue: Frequency of
  keyword ``Job hunting'' per one minute. Red: Alarm time.]{
   \includegraphics[width=.5\columnwidth]{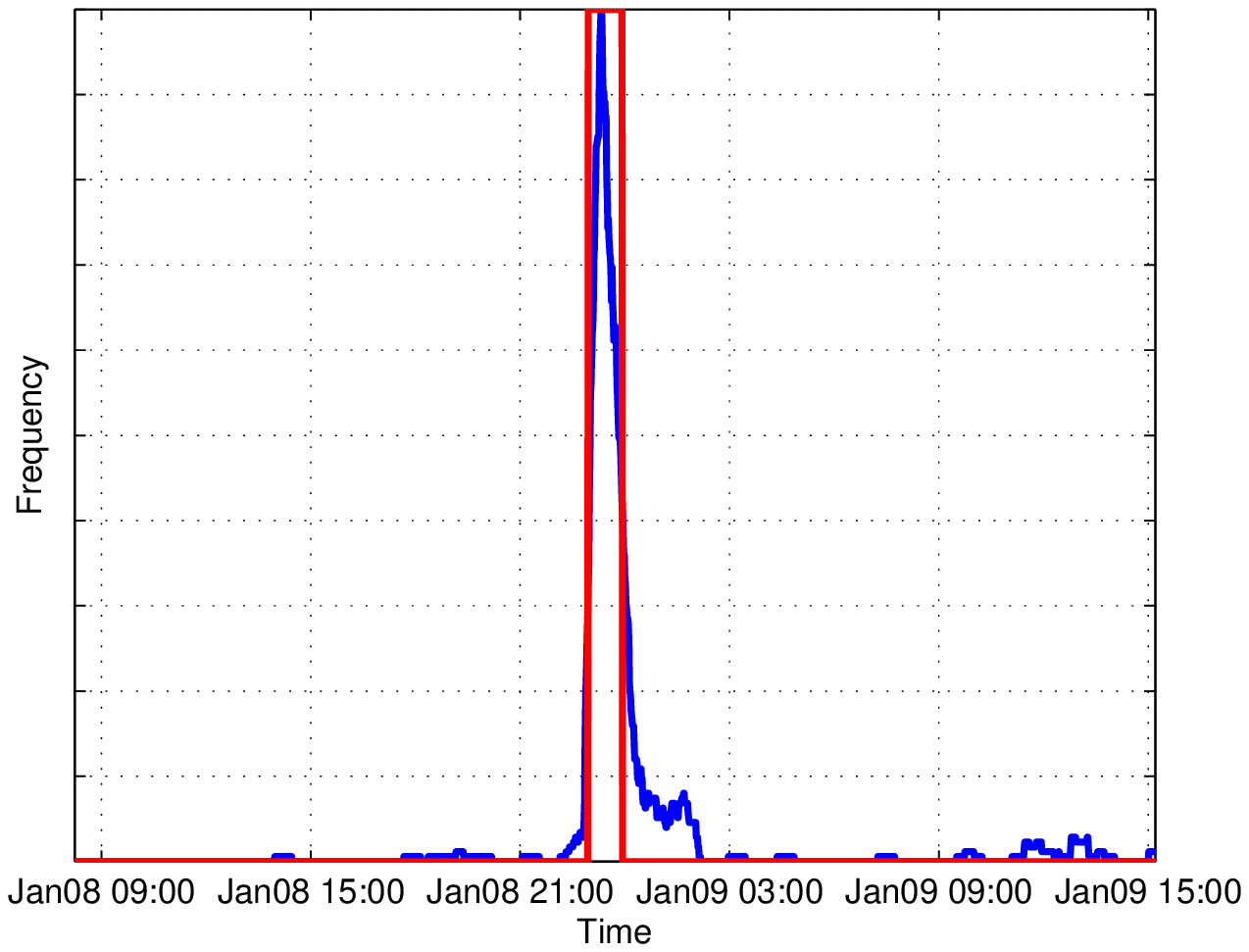}
  \label{fig:ikeda_word}}~\subfigure[Keyword-frequency-based burst detection. Blue: Frequency of
   keyword ``Job hunting'' per one second. Red: Burst state (burst or not).]{\includegraphics[width=.5\columnwidth]{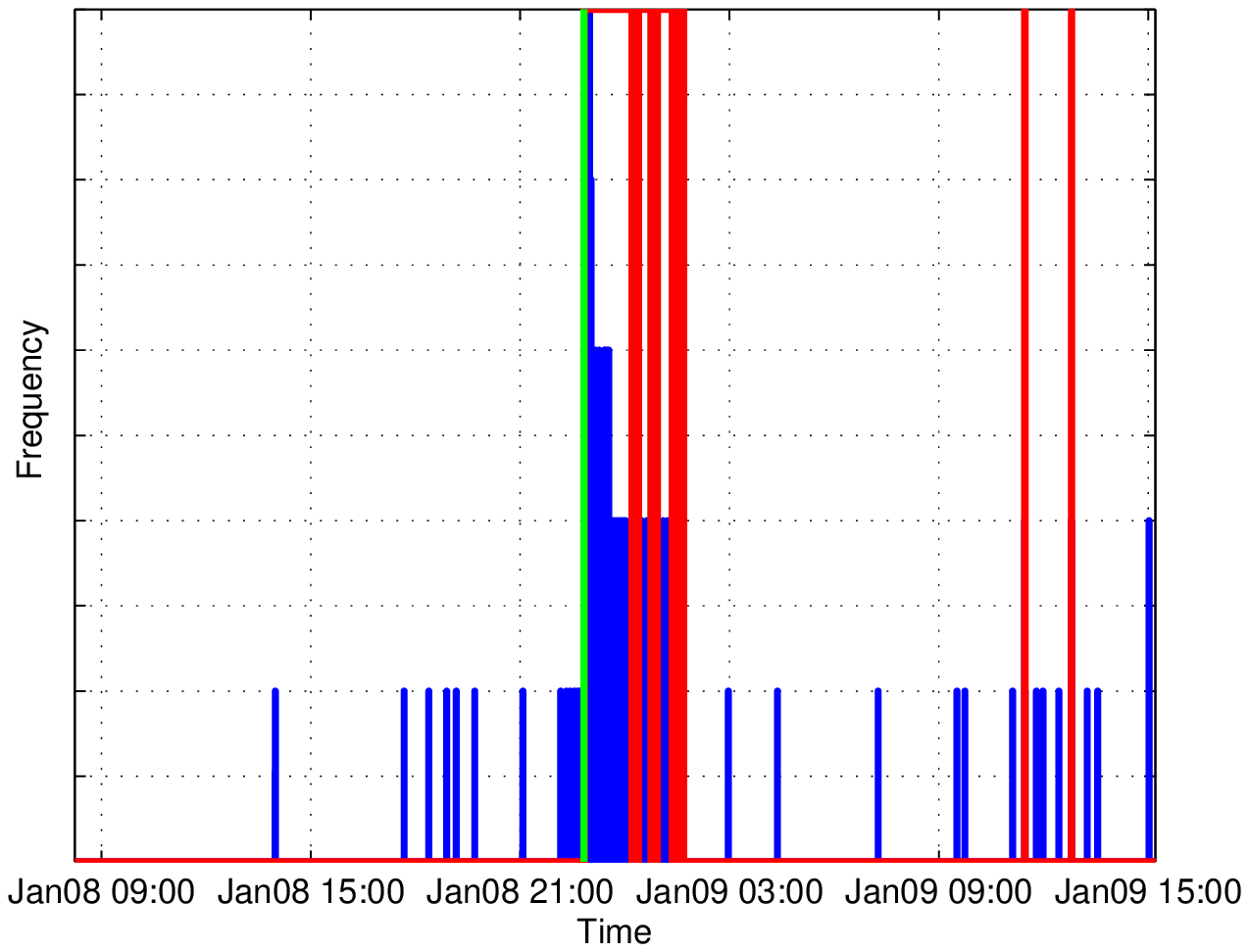}\label{fig:ikeda_word_burst}}
  \end{center}
 \caption{Result of ``Job hunting'' data set. The
initial controversial post was posted on 22:50, Jan 08 (green lines in
(b) and (d)).
} 
 \label{fig:ikeda-all}
\end{figure}

The keyword used in the keyword-based methods was ``Job hunting.''
Figures \ref{fig:ikeda} and \ref{fig:ikeda_link_burst} show the results
of the proposed link-anomaly-based change detection and burst
detection, respectively. Figures \ref{fig:ikeda_word} and 
\ref{fig:ikeda_word_burst} show the results of the keyword-frequency-based change
detection and burst detection, respectively.


The first alarm time of the proposed link-anomaly-based change-point analysis
was 22:55, whereas that for the keyword-frequency-based counterpart was
22:57; see also Table~\ref{tab:detection}. The earliest detection was achieved by the keyword-frequency-based
burst detection method. Nevertheless, from Figure~\ref{fig:ikeda-all}, we
can observe that the proposed link-anomaly-based methods were able to
detect the emerging topic almost as early as keyword-frequency-based methods.

\begin{table}[t]
 \begin{center}
 \caption{Detection time and the number of detections. The first
  detection time is defined as the time of the first alert after the
  event/post that initiated each topic; see captions for Figures
  \ref{fig:ikeda-all}--\ref{fig:bbc-all} for the details. }
  \label{tab:detection}
{\small
  \begin{tabular}[t]{|c|c|c|c|c|c|}
   \hline
Method & & ``Job hunting'' & ``Youtube'' & ``NASA'' & ``BBC'' \\\hline\hline
Link-anomaly-based   & $\sharp$ of detections   & 4     & 4 & 14 & 3 \\\cline{2-6}
 change-point detection & 1st detection time   &{\em 22:55, Jan 08}  & 08:44, Nov
	       05 &  {\em 20:11, Dec 02} & {\bf 19:52, Jan 21} \\\hline
Keyword-frequency-based     & $\sharp$ of detections   & 1     & 1 & 1 & 1\\\cline{2-6}
change-point detection & 1st detection time & 22:57, Jan 08 &  00:30, Nov 05 & 04:10,
		   Dec 03 & 22:41, Jan 21\\\hline
Link-anomaly-based   & $\sharp$ of detections   & 1     & 9          & 25 & 2 \\\cline{2-6}
burst detection & 1st detection time    & 23:07, Jan 08 & {\em 00:07, Nov
	       05} & {\bf 00:44, Nov 30} &
		   {\em 20:51, Jan 21} \\\hline
Keyword-frequency-based & $\sharp$ of detections & 6 & 15 & 11 & 1 \\\cline{2-6}
burst detection & 1st detection time & {\bf 22:50, Jan 08} & {\bf 23:59, Nov 04}
	       & 08:34, Dec 03 & 22:32, Jan 21 \\\hline
  \end{tabular}}
 \end{center}
 \label{fig:detect}
\end{table}

\subsection{``Youtube'' data set}
This data set is related to the recent leakage of some confidential video
by the Japan Coastal Guard officer.

\begin{figure}[tb]
  \begin{center}
 \subfigure[Link-anomaly-based change-point analysis. Green: Aggregated anomaly
  score~\eqref{eq:agg-s} at $\tau=1$ minute. Blue: Change-point
  score~\eqref{eq:score}. Red: Alarm time.]{
   \includegraphics[width=.5\columnwidth]{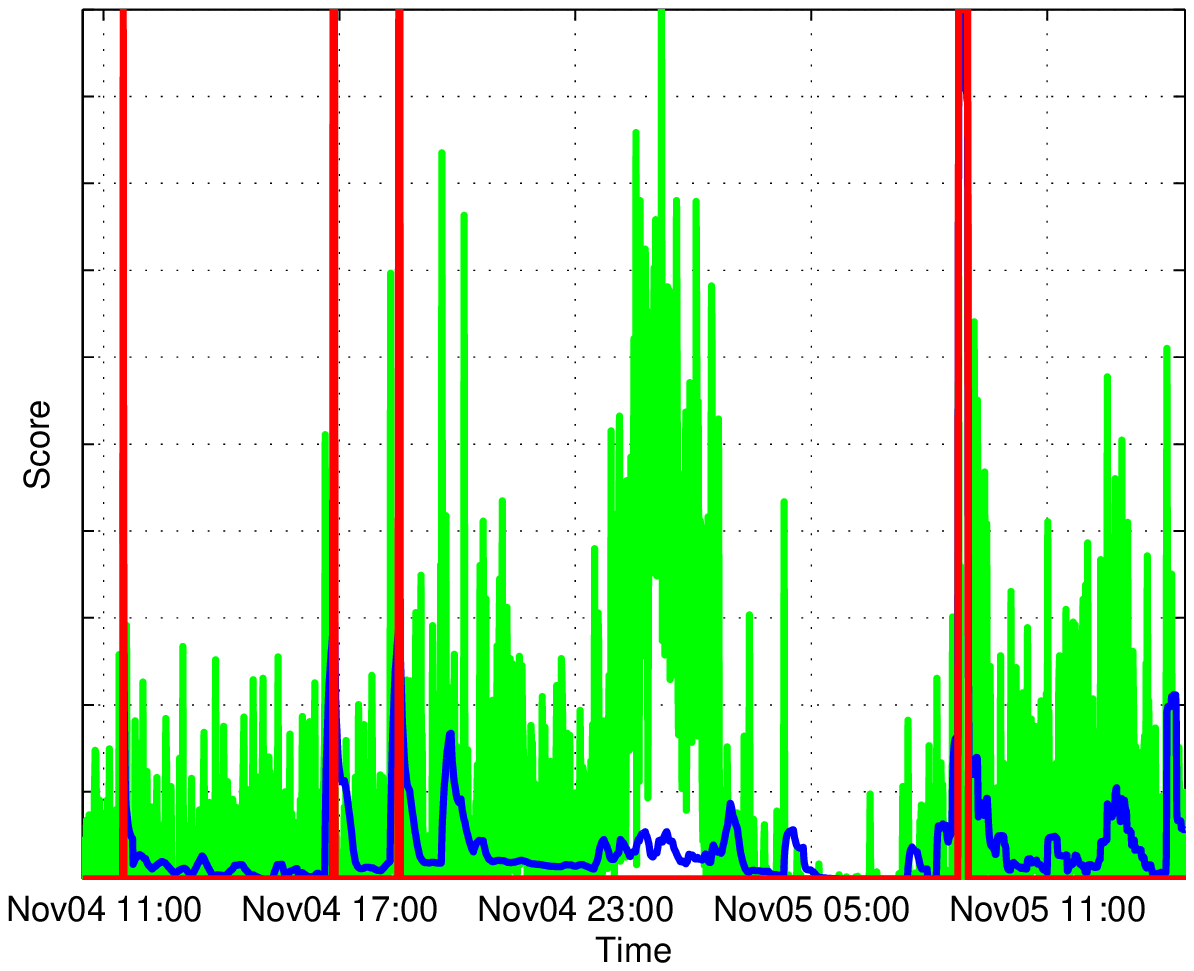}
\label{fig:senkaku}}~\subfigure[Link-anomaly-based burst
   detection.  Blue: Aggregated anomaly
  score~\eqref{eq:agg-s} at $\tau=1$ second.
 Cyan: threshold for the filtering step in Kleinberg's burst model.
 Red: Burst state.]{\includegraphics[width=.5\columnwidth]{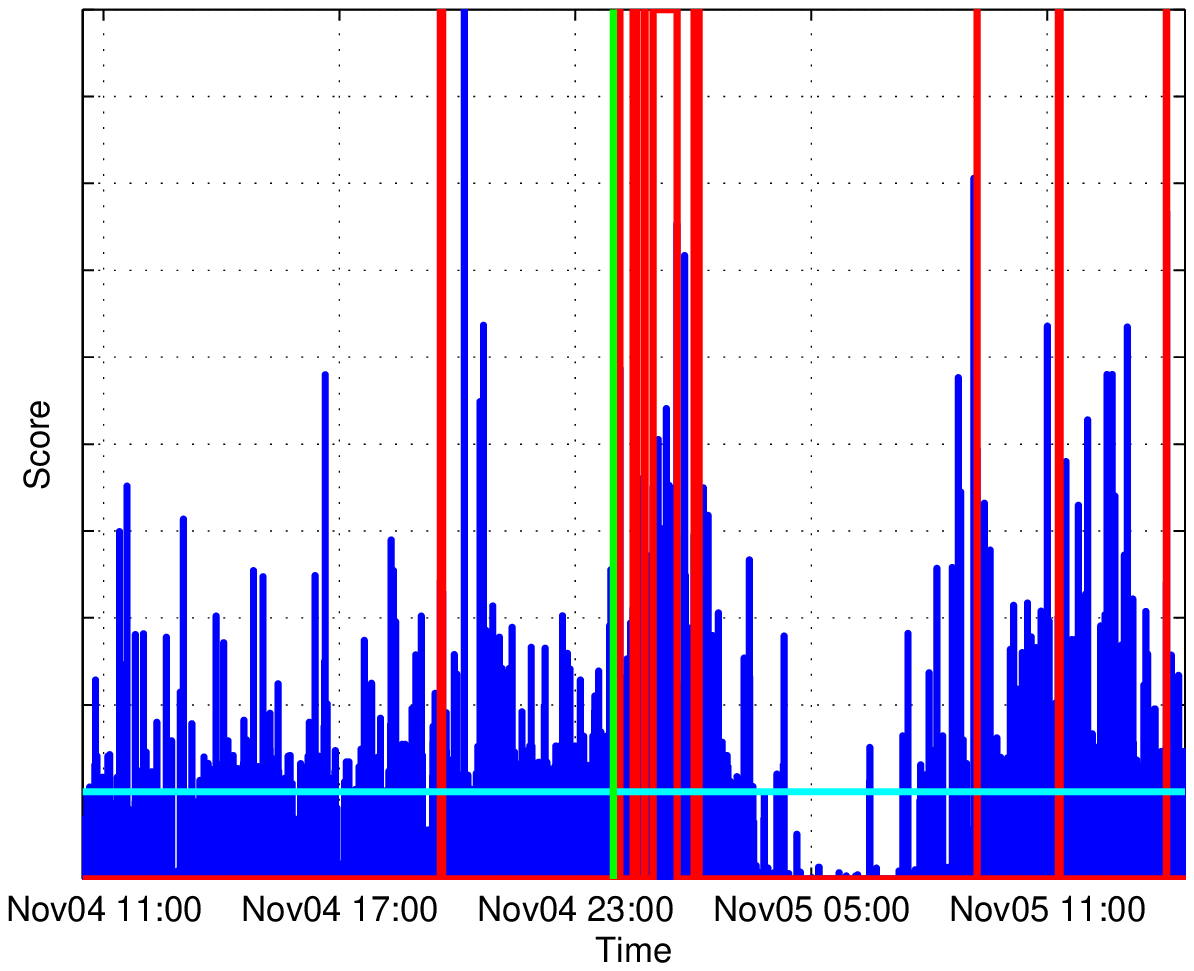}\label{fig:senkaku_link_burst}}
\subfigure[Keyword-frequency-based change-point analysis.  Blue: Frequency of
  keyword ``Senkaku'' per one minute. Red: Alarm time.]{
   \includegraphics[width=.5\columnwidth]{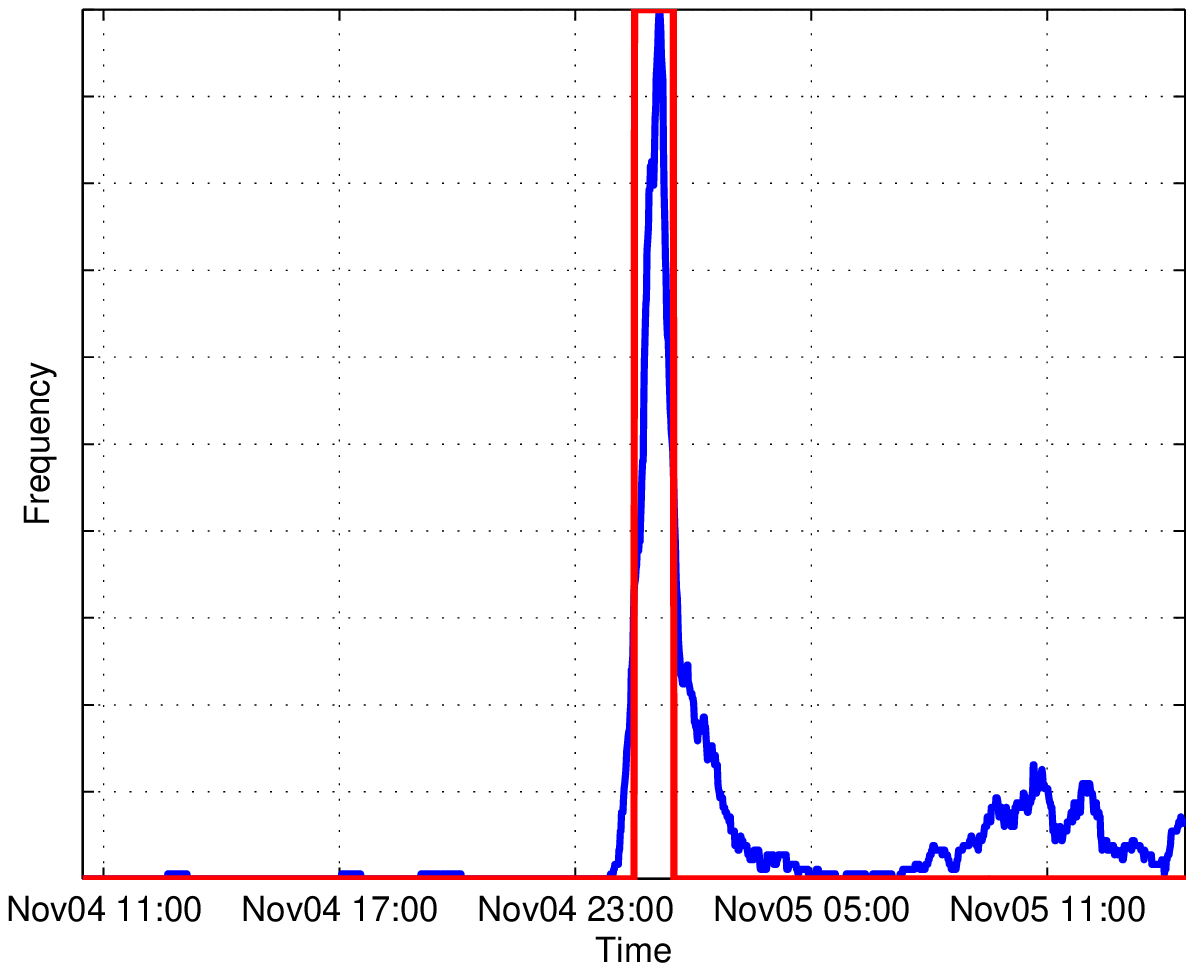}
  \label{fig:senkaku_word}}~\subfigure[Keyword-frequency-based burst detection.  Blue: Frequency of
   keyword ``Senkaku'' per one second. Red: Burst state (burst or not).]{\includegraphics[width=.5\columnwidth]{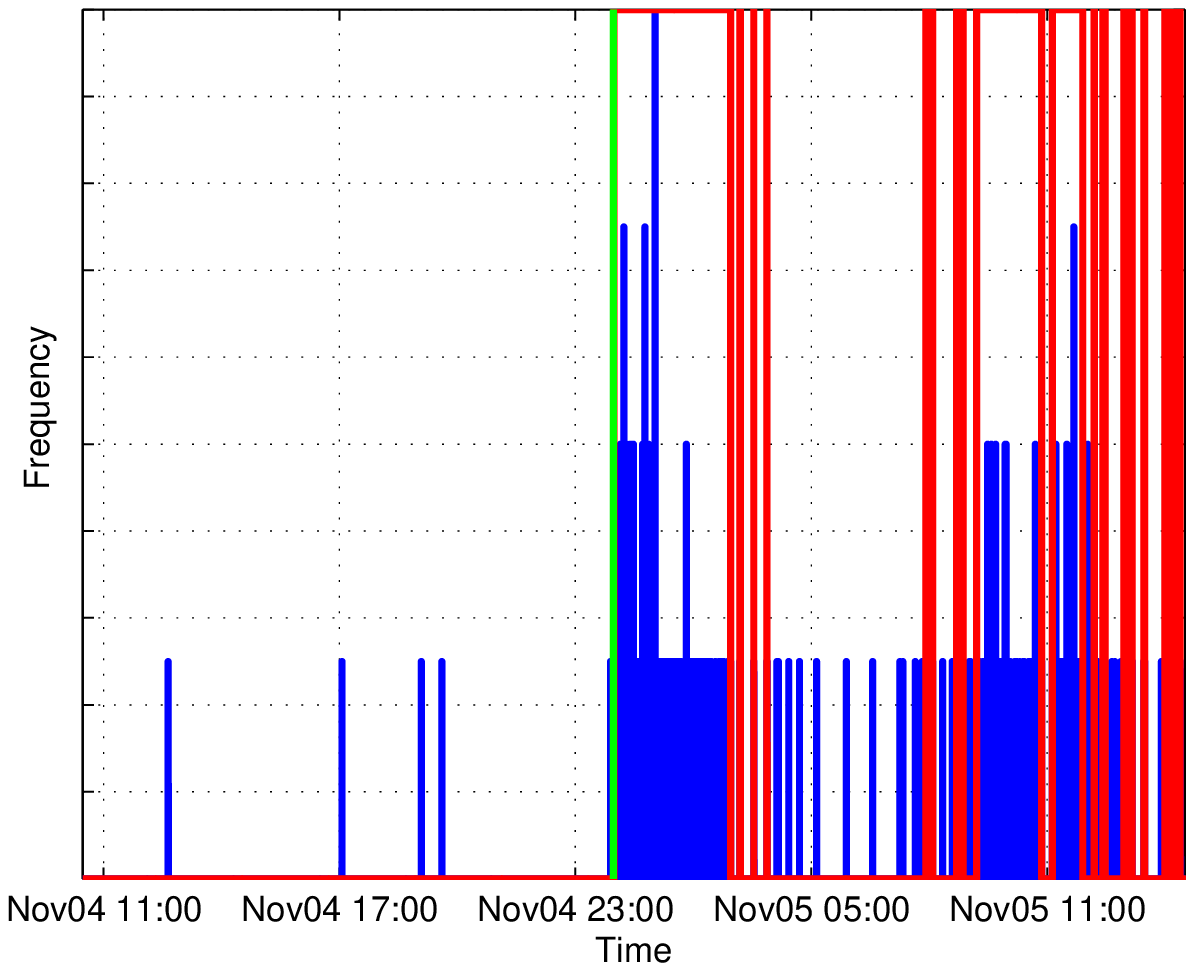}\label{fig:senkaku_word_burst}}
  \end{center}
 \caption{Result of ``Youtube'' data set. The first post about the
 video leakage was posted on 23:48, Nov 04 (green lines in (b) and (d)).}
 \label{fig:senkaku-all}
\end{figure}

The keyword used in the keyword-based methods is ``Senkaku.''
Figures \ref{fig:senkaku} and \ref{fig:senkaku_link_burst} show the results
of link-anomaly-based change detection and burst detection, respectively.
Figures \ref{fig:senkaku_word} and \ref{fig:senkaku_word_burst} show the
results of keyword-frequency based change detection and burst detection, respectively.

The first alarm time of the proposed link-anomaly-based change-point analysis
was 08:44, whereas that for the keyword-based counterpart was 00:30; see
also Table~\ref{tab:detection}. Although
the aggregated anomaly score~\eqref{eq:agg-s} in Figure
\ref{fig:senkaku} around midnight, Nov 05 is elevated, it seems that
SDNML fails to detect this elevation as a change point. In fact, 
the link-anomaly-based burst detection (Figure
\ref{fig:senkaku_link_burst}) raised an alarm at 00:07, which is earlier
than the keyword-frequency-based change-point analysis and closer to the
the keyword-frequency-based burst detection at 23:59, Nov 04.


\subsection{``NASA'' data set}
This data set is related to the discussion among Twitter users
interested in astronomy that preceded NASA's press conference about
discovery of an arsenic eating organism.

\begin{figure}[tb]
  \begin{center}
 \subfigure[Link-anomaly-based change-point analysis. Green: Aggregated anomaly
  score~\eqref{eq:agg-s} at $\tau=1$ minute. Blue: Change-point
  score~\eqref{eq:score}. Red: Alarm time.]{
   \includegraphics[width=.5\columnwidth]{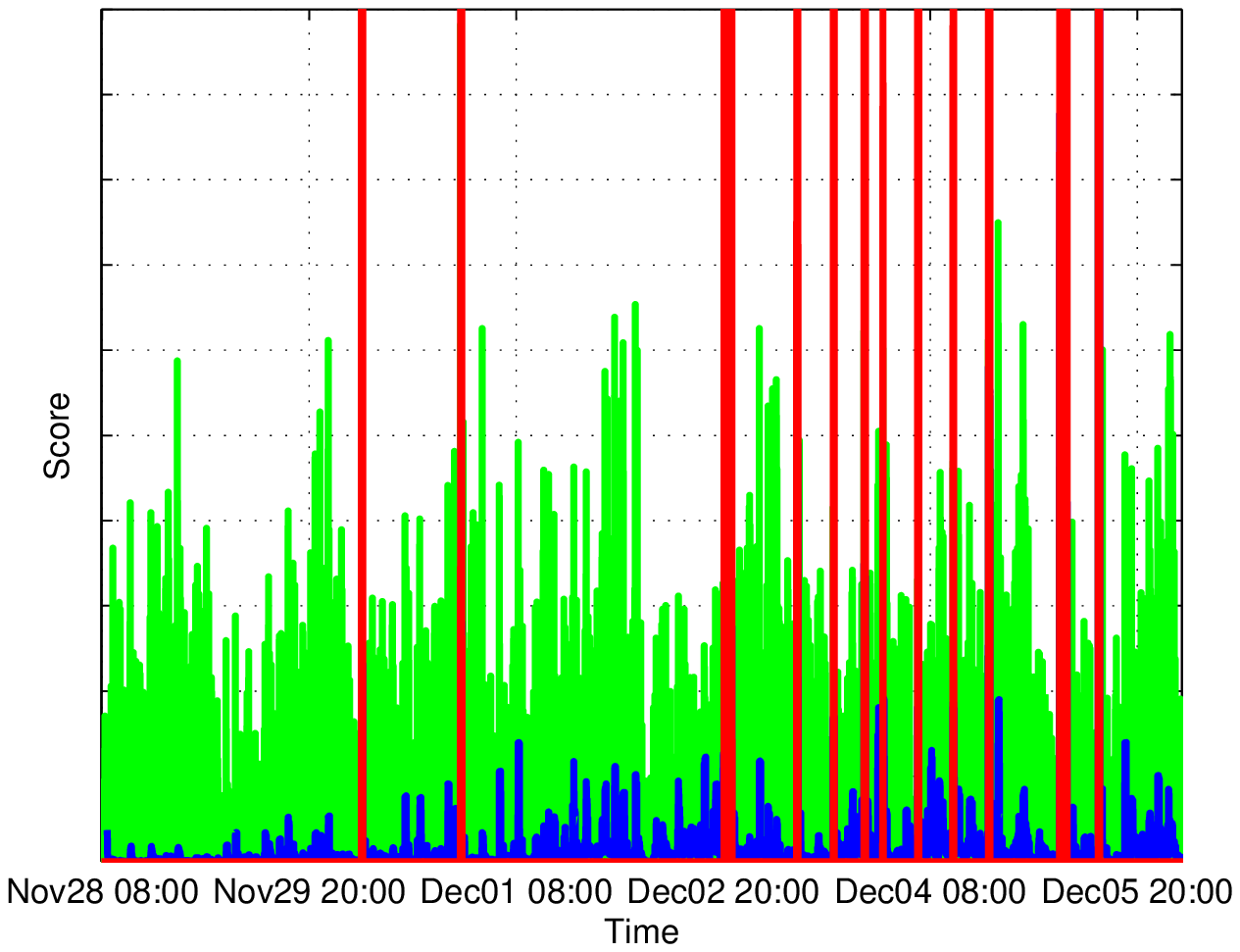}
\label{fig:nasa}}~\subfigure[Link-anomaly-based burst
   detection.  Blue: Aggregated anomaly
  score~\eqref{eq:agg-s} at $\tau=1$ second.
 Cyan: threshold for the filtering step in Kleinberg's burst model.
 Red: Burst state.]{\includegraphics[width=.5\columnwidth]{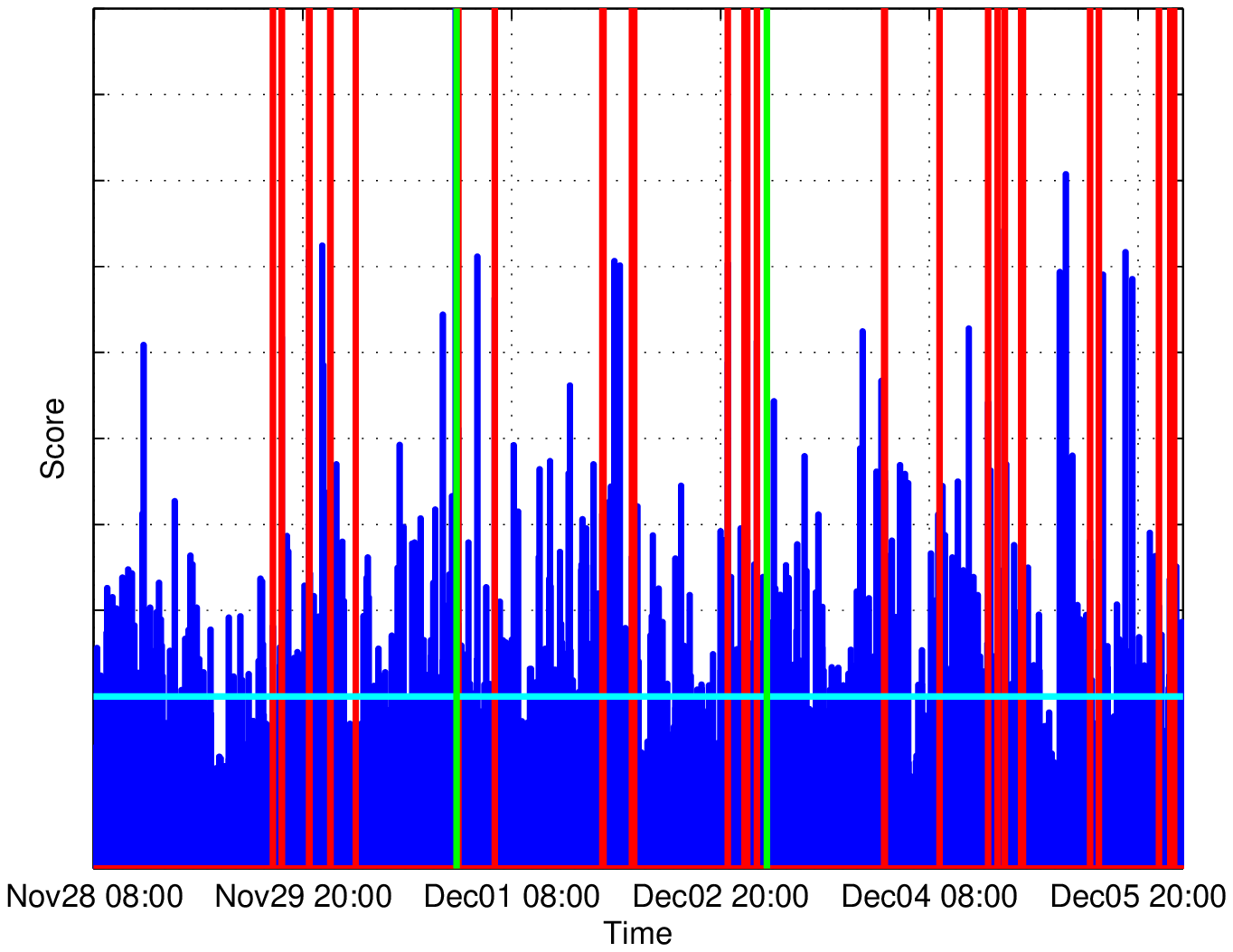}\label{fig:nasa_link_burst}}
\subfigure[Keyword-frequency-based change-point analysis.  Blue: Frequency of
  keyword ``arsenic'' per one minute. Red: Alarm time.]{
   \includegraphics[width=.5\columnwidth]{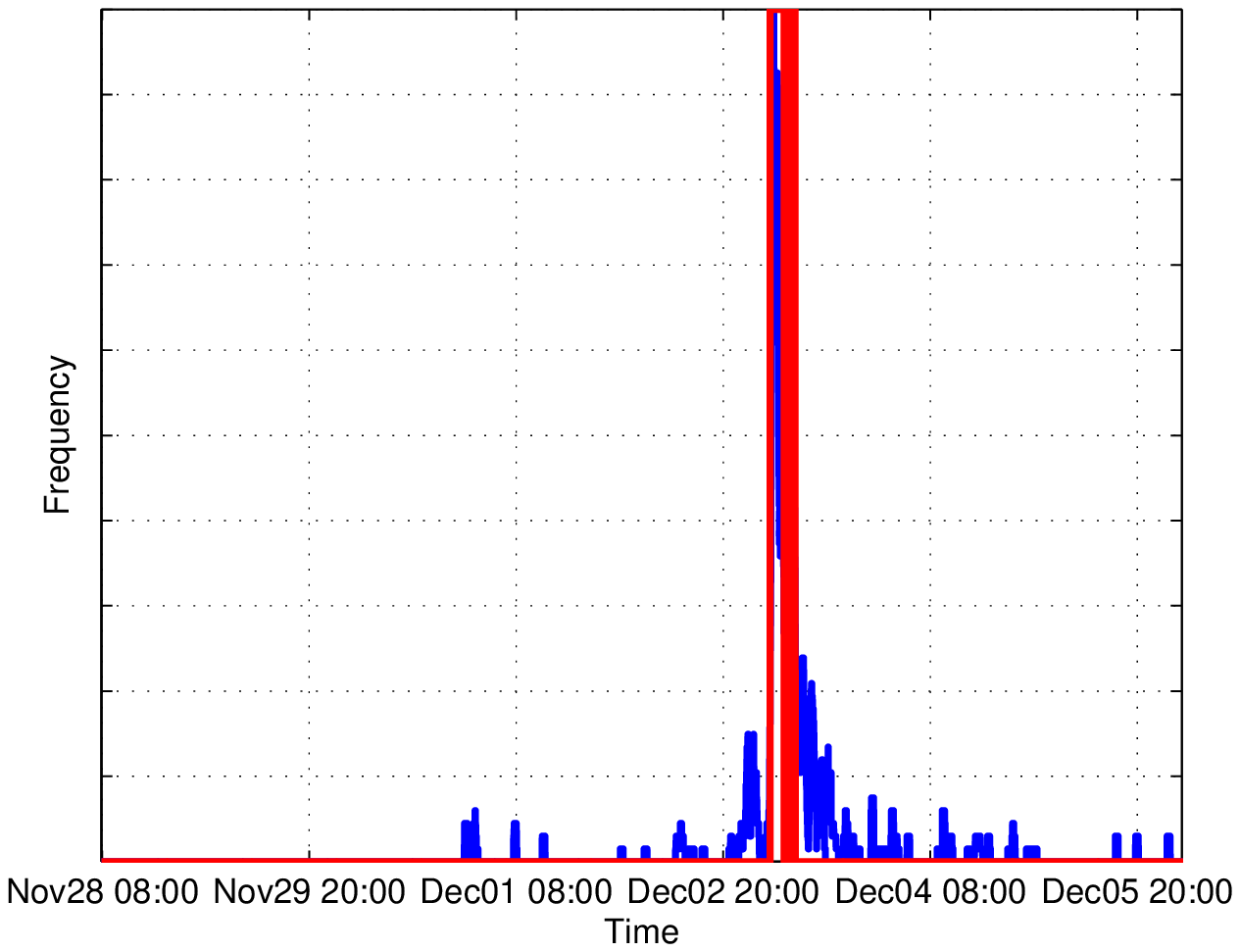}
  \label{fig:nasa_word}}~\subfigure[Keyword-frequency-based burst detection.  Blue: Frequency of
   keyword ``arsenic'' per one second. Red: Burst state (burst or not).]{\includegraphics[width=.5\columnwidth]{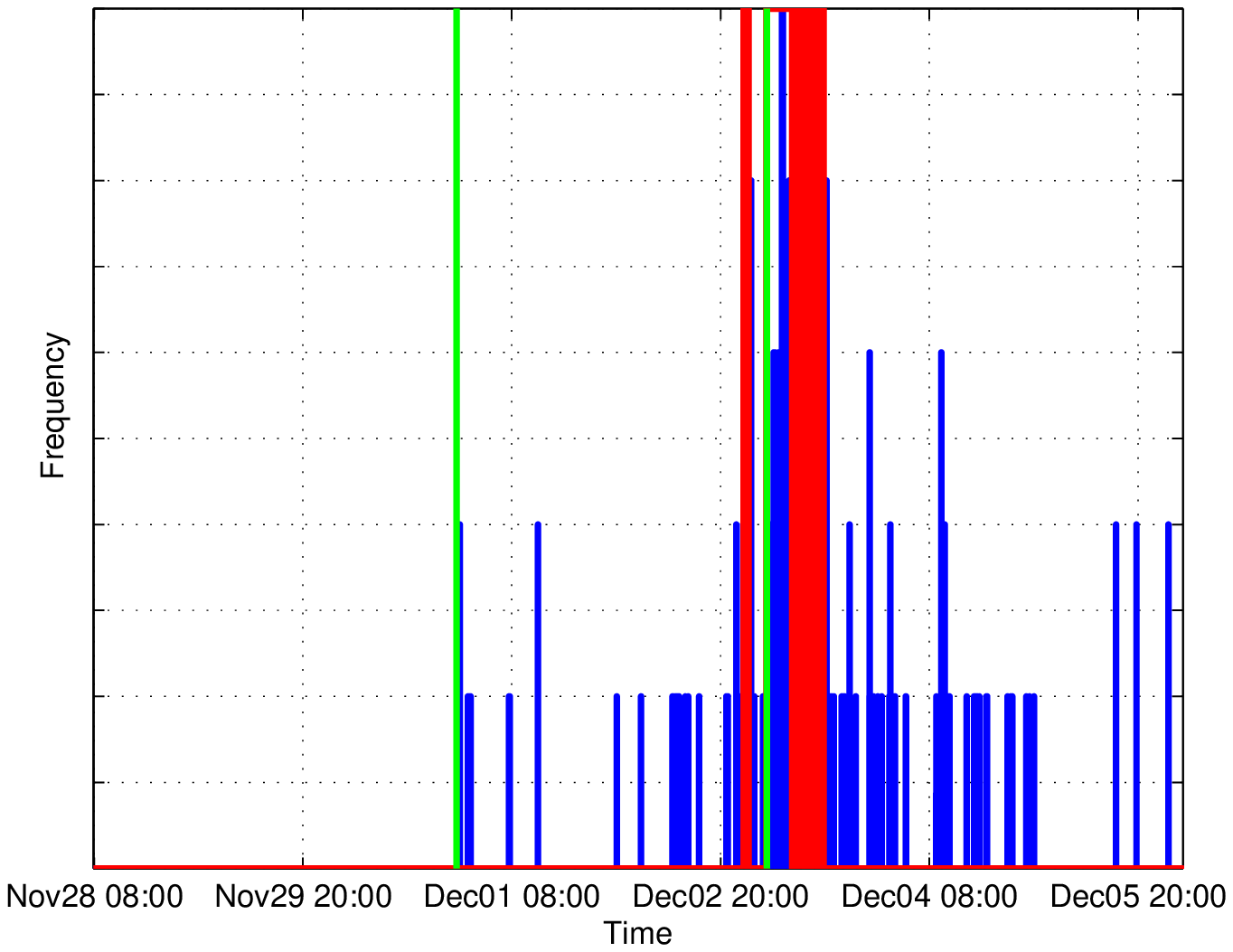}\label{fig:nasa_word_burst}}
  \end{center}
 \caption{Result of ``NASA'' data set. The initial post predicting
 NASA's finding about arsenic-eating organism was posted on 22:30, Nov
 30 much earlier than NASA's official press conference at 04:00, Dec 03. }
\end{figure}

The keyword used in the keyword-based models is ``arsenic.''
Figures \ref{fig:nasa} and \ref{fig:nasa_link_burst} show the results of
link-anomaly-based change detection and burst detection, respectively.
Figures \ref{fig:nasa_word} and \ref{fig:nasa_word_burst} show the same
results for the keyword-frequency-based methods.

The first alarm times of the two link-anomaly-based methods were 20:11,
Dec 02 (change-point detection) and 00:44, Nov 30 (burst detection),
respectively. Both of these are earlier than NASA's official press
conference (04:00, Dec 03) and are earlier than the keyword-frequency
based methods (change-point detection at 04:10, Dec
03 and burst detection at 08:34, Dec 03.); see Table~\ref{tab:detection}.

\subsection{''BBC'' data set}
This data set is related to angry reactions among Japanese Twitter users
against a BBC comedy show that asked ``who is the unluckiest person in
the world'' (the answer is a Japanese man who got hit by nuclear bombs
in both Hiroshima and Nagasaki but survived).

\begin{figure}[tb]
  \begin{center}
 \subfigure[Link-anomaly-based change-point analysis. Green: Aggregated anomaly
  score~\eqref{eq:agg-s} at $\tau=1$ minute. Blue: Change-point
  score~\eqref{eq:score}. Red: Alarm time.]{
   \includegraphics[width=.5\columnwidth]{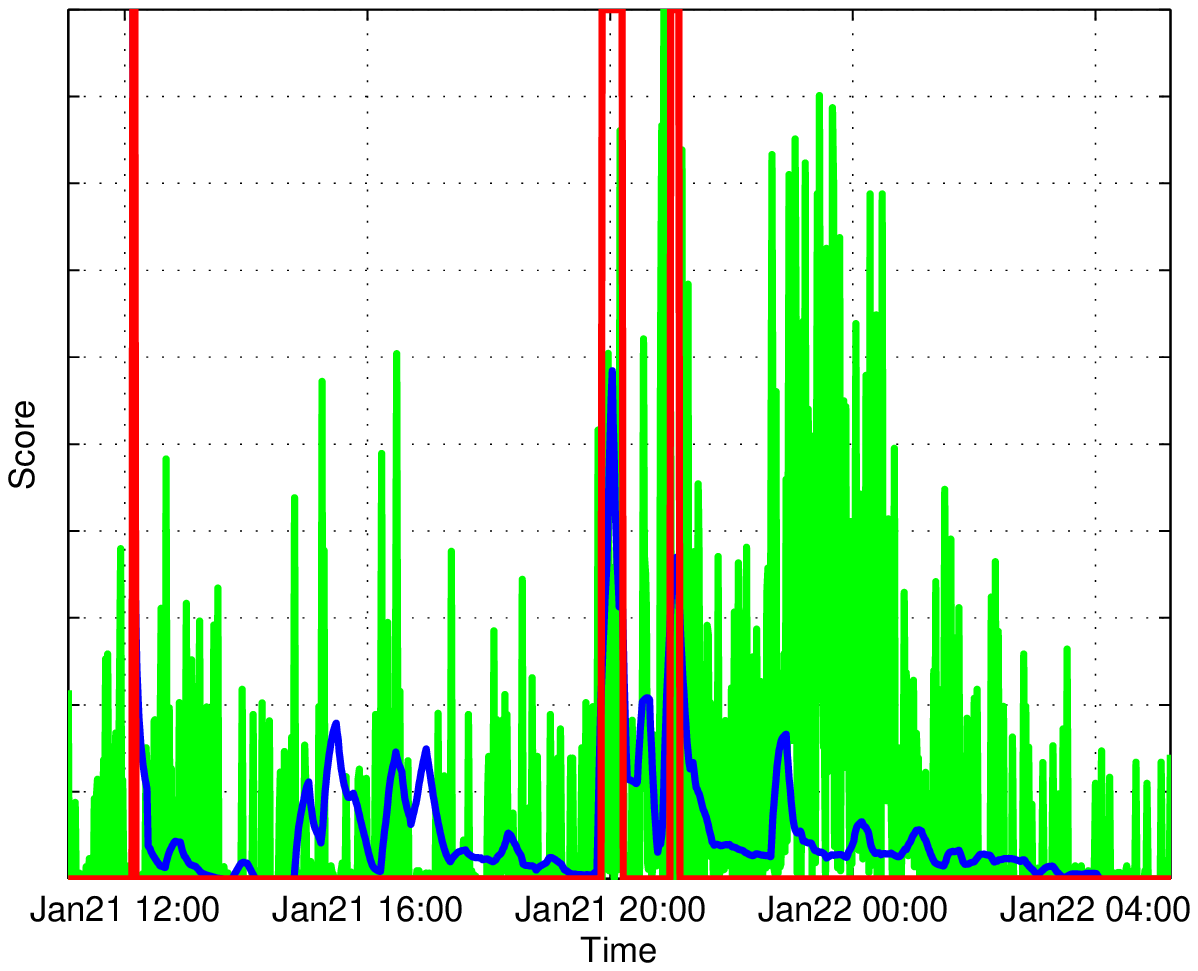}
\label{fig:bbc}}~\subfigure[Link-anomaly-based burst
   detection.  Blue: Aggregated anomaly
  score~\eqref{eq:agg-s} at $\tau=1$ second.
 Cyan: threshold for the filtering step in Kleinberg's burst model.
 Red: Burst state.]{\includegraphics[width=.5\columnwidth]{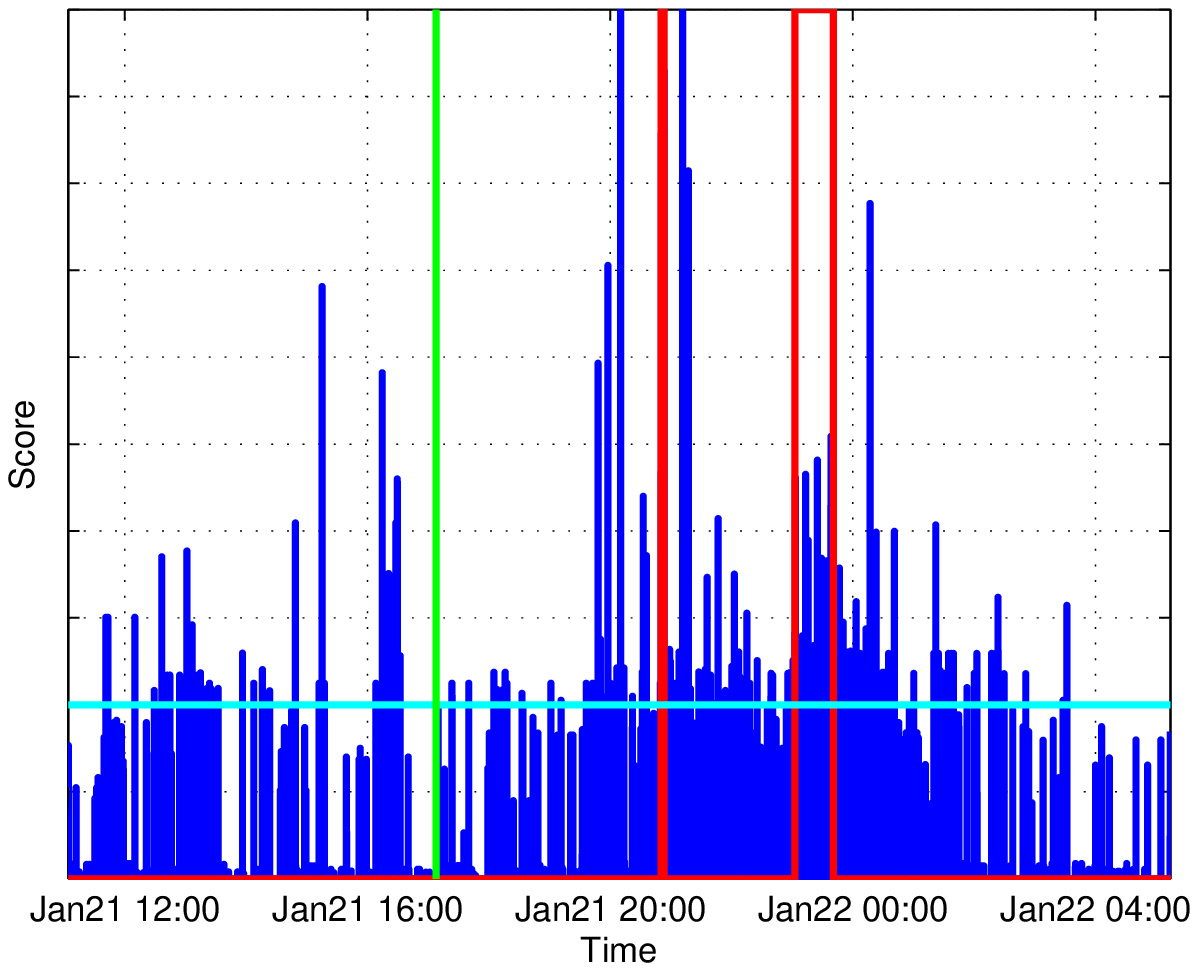}\label{fig:bbc_link_burst}}
   \subfigure[Keyword-frequency-based change-point analysis.  Blue: Frequency of
  keyword ``British'' per one minute. Red: Alarm time.]{
   \includegraphics[width=.5\columnwidth]{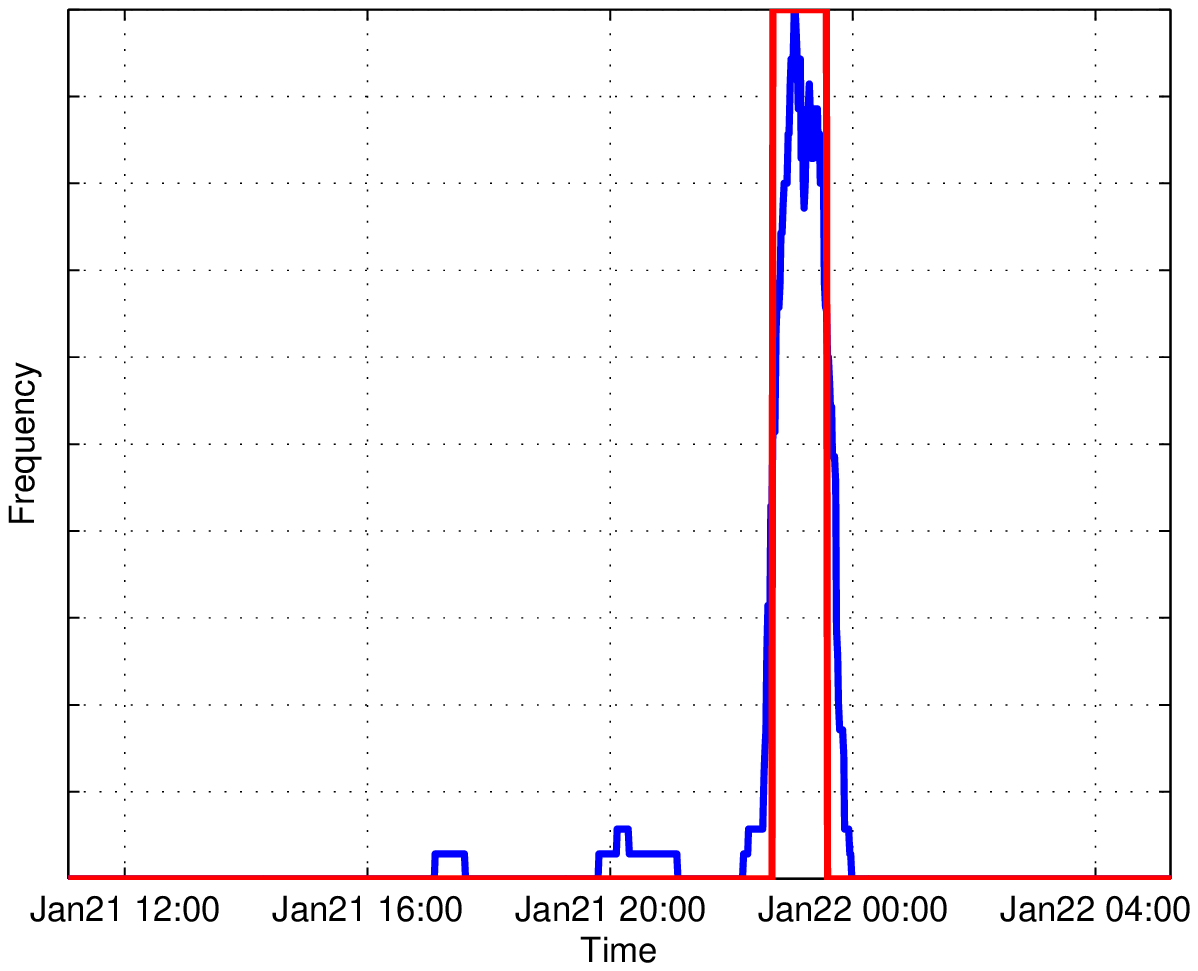}
  \label{fig:bbc_word}}~\subfigure[Keyword-frequency-based burst detection.  Blue: Frequency of
   keyword ``British'' per one second. Red: Burst state (burst or not).]{\includegraphics[width=.5\columnwidth]{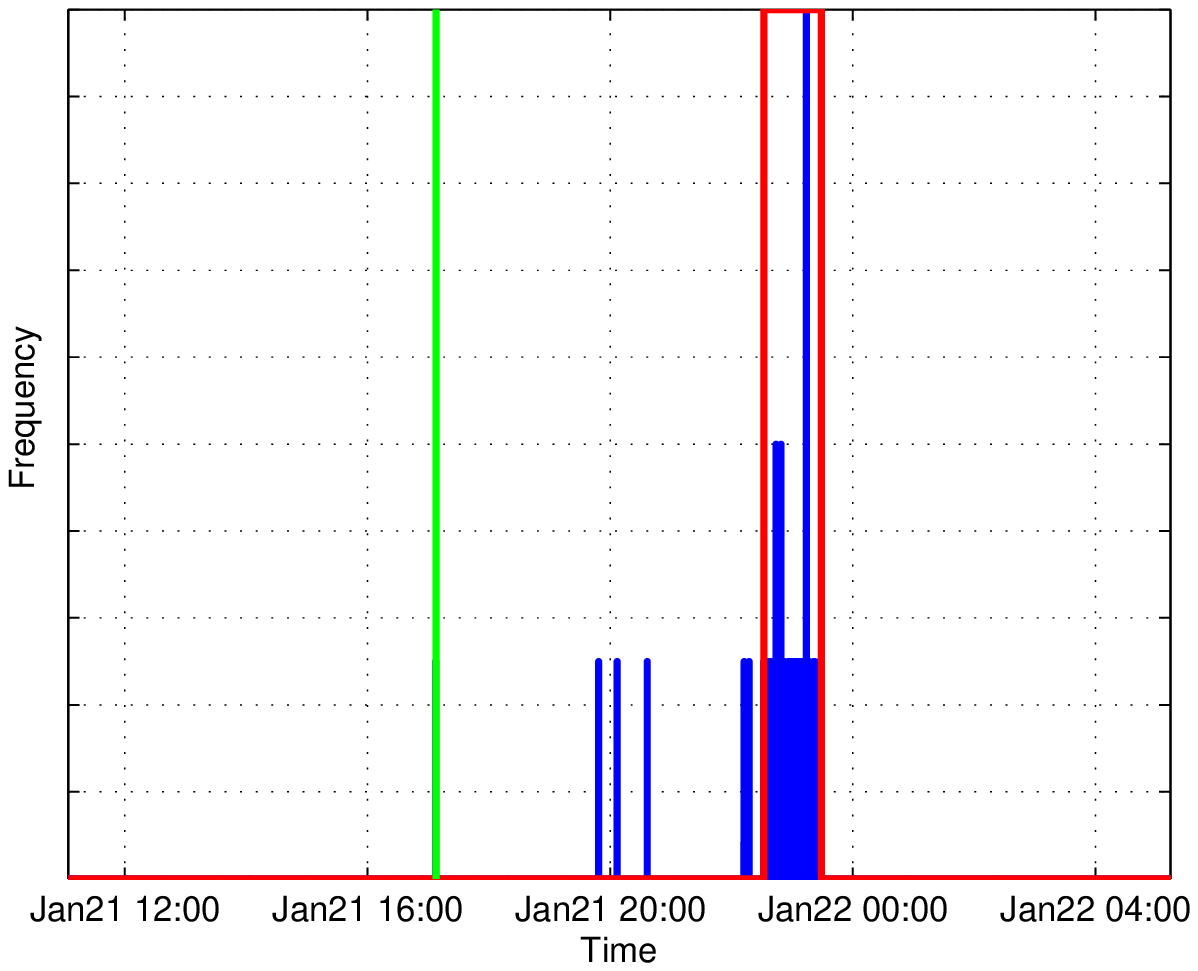}\label{fig:bbc_word_burst}}
  \end{center}
 \caption{Result of ``BBC'' data set. The first post about BBC's comedy
 show was posted on 17:08, Jan 21 (green lines in (b) and (d)).}
 \label{fig:bbc-all}
\end{figure}

The keyword used in the keyword-based models is ``British'' (or ``Britain'').
Figures \ref{fig:bbc} and \ref{fig:bbc_link_burst} show the results of
link-anomaly-based change detection and burst detection, respectively.
Figures \ref{fig:bbc_word} and \ref{fig:bbc_word_burst} show the same
results for the keyword-frequency-based methods.

The first alarm time of the two link-anomaly-based methods 
was 19:52 (change-point detection) and 20:51 (burst detection), both of which
are earlier than the keyword-frequency-based counterparts at 22:41
(change-point detection) and 22:32 (burst detection). See Table~\ref{tab:detection}.

\subsection{Discussion}
Within the four data sets we have analyzed above, the proposed
link-anomaly based methods compared favorably against the keyword-frequency
based methods on ``NASA'' and ``BBC'' data sets. On the other hand, the
keyword-frequency based methods were earlier to detect the topics on ``Job hunting'' and
``Youtube'' data sets.

The above observation is natural, because for ``Job hunting'' and
``Youtube'' data sets, the keywords seemed to have been unambiguously
defined from the beginning of the emergence of the topics, whereas for
``NASA'' and ``BBC'' data sets, the keywords are more ambiguous. In
particular, in the case of ``NASA'' data set, people had been mentioning
``arsenic'' eating organism {\em earlier} than NASA's official release but only rarely (see Figure~\ref{fig:nasa_word_burst}). Thus, the keyword-frequency-based methods could
not detect the keyword as an emerging topic, although the keyword
``arsenic'' appeared earlier than the official release. For ``BBC''
data set, the proposed link-anomaly-based burst model detects two bursty
areas (Figure~\ref{fig:bbc_link_burst}). Interestingly, the link-anomaly-based change-point analysis only
finds the first area (Figure~\ref{fig:bbc}), whereas the
keyword-frequency-based methods only find the second area
(Figures~\ref{fig:bbc_word} and \ref{fig:bbc_word_burst}).
This is probably because there was an initial stage where people reacted
individually using different words and later there was another stage in
which the keywords are more unified.

In our approach, the alarm was raised if the change-point score exceeded a
dynamically optimized threshold  based on the significance level
parameter $\rho$. 
Table \ref{fig:proposed_detect} shows results for a number of  threshold parameter values.
We see that as $\rho$ increased, the number of false alarms also increased.
Meanwhile, even when it was so small, our approach was still able to
detect the emerging topics as early as the keyword-based methods.
We set $\rho =0.05$ as a default parameter value in our experiment.
Although there are several alarms for ``NASA'' data set, most of them are
more or less related to the emerging topic.





\begin{table}[t]
 \begin{center}
 \caption{Number of alarms for the proposed change-point detection
 method based on the link-anomaly score~\eqref{eq:agg-s} for various
  significance level parameter values $\rho$.}  
 \label{fig:proposed_detect}
  \begin{tabular}[t]{|c|c|c|c|c|}
   \hline
   $\rho$&``Job hunting''&``Youtube''&``NASA'' & ``BBC''\\\hline\hline
   0.01& 4 &  2    & 9       &  3 \\\hline
   0.05& 4 &  4    & 14       &  3 \\\hline
   0.1&  8 &  6    & 30     &  3 \\\hline
  \end{tabular}
 \end{center}
\end{table}

Notice again that in the keyword-based methods the keyword related to
the topic must be known in advance, which is not always the case in
practice.
Further note that our approach only uses links (mentions), hence it can
be applied to the case where topics are concerned with  information
other than texts, such as images, video, sounds, etc. 
 %


\section{Conclusion}
In this paper, we have proposed a new approach to detect the emergence
of topics in a social network stream. The basic idea of our approach is
to focus on the social aspect of the posts reflected in the mentioning
behaviour of users instead of the textual contents. We have proposed a
probability model that captures both the number of mentions per post and
the frequency of mentionee. We have combined the proposed mention model with
the SDNML change-point detection algorithm~\cite{UraYamTomIwa11} and
Kleinberg's burst detection model~\cite{Kle03} to pin-point the emergence of a topic.

We have applied the proposed approach to four real data sets we have
collected from Twitter. The four data sets included a wide-spread
discussion about a controversial topic (``Job hunting'' data set), a
quick propagation of news about a video leaked on Youtube (``Youtube''
data set), a rumor about the upcoming press conference by NASA
(``NASA'' data set), and an angry response to a foreign TV show (``BBC''
data set).  In all the data sets our proposed approach showed
promising performance. In most data set,  the detection by the proposed
approach was as early as term-frequency based approaches in the
hindsight of the keyword that best describes the topic that we have
manually chosen afterwards. Furthermore, for ``NASA'' and ``BBC'' data
sets, in which the keyword that defines the topic is more ambiguous than
the first two data sets, the proposed link-anomaly based approaches
have detected the emergence of the topics much earlier than the
 keyword-based approaches.

All the analysis presented in this paper was conducted off-line but the
framework itself can be applied on-line. We are planning to scale up the
proposed approach to handle social streams in real time. It would also be
interesting to combine the proposed link-anomaly model with
content-based topic detection approaches to further boost the performance and reduce
false alarms.

%
%

\section*{Acknowledgments}
This work was partially supported by MEXT KAKENHI 23240019, 22700138,
Aihara Project, the FIRST program from JSPS, initiated by CSTP, Hakuhodo
Corporation, NTT Corporation, and Microsoft Corporation (CORE Project).

%
%
%



\appendix
\section{Sequentially discounting normalized maximum likelihood coding}
\label{sec:sdnml}
This section describes the sequentially discounting normalized maximum
likelihood (SDNML) coding that we use for change-point detection in
Section~\ref{sec:change-point}. The basic idea behind SDNML-based change
detection is as follows: when the data arrives in a sequential
manner, we can consider a change has occurred if a new piece of
data cannot be compressed using the statistical nature of the past. The
original paper~\cite{YamTak02,TakYam06} used the predictive
stochastic complexity as a measure of compressibility, whereas Urabe et
al. \cite{UraYamTomIwa11} proposed to employ a tighter coding scheme
based on the SDNML.

Suppose that we observe a discrete time series $x_t$
 ($t=1,2,\ldots,$); we denote the data sequence by $x^t:=x_1\cdots x_t$. Consider the
 parametric class of conditional probability densities
 $\mathcal{F}=\{p(x_t|x^{t-1}:\vtheta):\vtheta\in\RR^p\}$, where $\vtheta$ is the
 $p$-dimensional parameter vector and we assume $x^{0}$ to be an empty
 set. We denote the maximum likelihood (ML) estimator given the data sequence
 $x^t$ by  $\hat{\vtheta}(x^t)$; i.e.,  $\hat{\vtheta}(x^t):={\rm
 argmax}_{\vtheta\in\RR^p}\prod_{j=1}^{t}p(x_j|x^{j-1}:\vtheta)$. 
 The sequential normalized maximum likelihood (SNML) model is a coding
distribution (see e.g., \cite{CovTho91}) that is known to be optimal in the
sense of the conditional minimax~\cite{RooRis08} problem:
\begin{align}
\label{eq:minimax}
\hspace*{-5mm} \min_{q(\cdot|x^{t-1})}\max_{x_t}\left(-\log
 q(x_t|x^{t-1})+\max_{\vtheta\in\RR^p}\log p(x^t:\vtheta)\right),
\end{align}
where $p(x^t):=\prod_{j=1}^{t}p(x_j|x^{j-1}:\vtheta)$ is the joint
density over $x^t$ induced by the conditional densities from $\mathcal{F}$.
The minimization is taken over all conditional density functions and
tries to minimize the regret~\eqref{eq:minimax} over any possible
outcome of the new sample $x_t$.

The SNML distribution is obtained as the optimal conditional density of
the minimax problem~\eqref{eq:minimax} as follows~\cite{RooRis08}:
\begin{align}
 \label{eq:snml}
p_{\mbox{\tiny SNML}}(x_t|x^{t-1}):=\frac{p(x^t:\hat{\vtheta}(x^t))}{K_t(x^{t-1})},
\end{align}
where the normalization constant $K_t(x^{t-1}):=\int
p(x^t|\hat{\vtheta}(x^t)){\rm d} x_t$ is necessary because the new sample
$x_t$ is used in the estimation of parameter vector $\hat{\vtheta}(x^t)$
and the numerator in \eqref{eq:snml} is not a proper density function.
We call the quantity  $-\log p_{\mbox{\tiny SNML}}(x_{t}|x^{t-1})$ the {\em SNML code-length}.
It is known from \cite{RooRis08,RisRooMyl10} that the cumulative SNML
code-length,  which is the sum of SNML code-length over the sequence, is optimal 
in the sense that  it asymptotically achieves the shortest code-length.

The sequentially discounting normalized maximum likelihood (SDNML) is
obtained by applying the above SNML to the class of autoregressive (AR) model
and replacing the ML estimation in \eqref{eq:snml} with a {\em
discounted} ML estimation, which makes
the SDNML-based change-point detection algorithm more flexible than an
SNML-based one. Let $x_{t}\in\RR$
for each $t$. We define the $p$th order AR model as follows:
\begin{align*}
p(x_t|x_{t-k}^{t-1}:\vtheta)=\frac{1}{\sqrt{2\pi\sigma^2}}\exp\Biggl(
-\frac{1}{2\sigma^2}\Bigl(x_t-\sum_{i=1}^{p}a^{(i)}x_{t-i}\Bigr)^2\Biggr),
\end{align*}
where $\vtheta\T=(\va\T,\sigma^2)=((a^{(1)},\ldots,a^{(p)}),\sigma^2)$ is the parameter vector.

In order to compute the SDNML density function we need the
discounted ML estimators of the parameters in $\vtheta$. We define the
discounted ML estimator of the regression coefficient $\hat{\va}_t$ as
follows:
\begin{align}
\label{eq:discount-a}
 \hat{\va}_t&=\argmin_{\va\in\RR^{p}}\sum_{j=t_0+1}^{t}w_{t-j}\left(x_j-\va\T\bar{\vx}_j\right)^2,
\end{align}
where $w_{j'}=r(1-r)^{j'}$ is a sequence of sample weights with the
discounting coefficient $r$ ($0<r<1$); $t_0$ is the smallest
number of samples such that the minimizer~\eqref{eq:discount-a} is
unique; $\bar{\vx}_j:=(x_{j-1},x_{j-2},\ldots,x_{j-k})\T$. Note that
the error terms from older samples receive geometrically decreasing weights
in \eqref{eq:discount-a}. The larger the discounting coefficient $r$ is,
the smaller the weights of the older samples become; thus we have
stronger discounting effect. Moreover, we obtain the discounted ML
estimator of the variance $\hat{\tau}_t$ as follows:
\begin{align*}
\hat{\tau}_t:=&\argmax_{\sigma^2}\prod_{j=t_0+1}^{t} p\left(x_j|x_{j-k}^{j-1}:\hat{\va}_j,\sigma^2\right)\\
=&\frac{1}{t-t_{0}}\sum_{j=t_0+1}^{t}\hat{e}_j^2=\frac{S_t}{t-t_0},
\end{align*}
where we define
$\hat{e}_j^2=\left(x_j-\hat{\va}_j\T\bar{\vx}_j\right)^2$ and $S_t:=\sum_{j=t_0+1}^{t}\hat{e}_j^2$.
Clearly when the discounted estimator of the AR coefficient $\hat{\va}_j$
is available, $S_t$ can be computed in a sequential manner.

In the sequel, we first describe how to efficiently compute the
AR estimator $\hat{\va}_j$. Finally we derive the SDNML density function
using  the discounted ML estimators
$(\hat{\va}_t,\hat{\tau}_t)$.

The AR coefficient $\hat{\va}_j$ can simply be computed by solving the
least-squares problem~\eqref{eq:discount-a}. It can, however, be
obtained more efficiently using the iterative formula described in
\cite{RooRis08,RisRooMyl10}. Here we repeat the formula for the
discounted version presented in \cite{UraYamTomIwa11}. First define the sufficient statistics $V_t\in\RR^{p\times p}$ and $\vchi_t\in\RR^{p}$ as follows:
\begin{align*}
 V_{t}     :=\sum_{j=t_0+1}^{t}w_j\bar{\vx}_j\bar{\vx}_j\T, 
\quad \vchi_{t} :=\sum_{j=t_0+1}^{t}w_j\bar{\vx}_jx_j.
\end{align*}
Using the sufficient statistics, the discounted AR coefficient
$\hat{\va}_j$ from \eqref{eq:discount-a} can be written as follows:
\begin{align*}
 \hat{\va}_{t}&=V_{t}^{-1}\vchi_{t}.
\end{align*}
Note that $\vchi_{t}$ can be computed in a sequential manner. The inverse
matrix $V_{t}^{-1}$ can also be computed sequentially using the
Sherman-Morrison-Woodbury formula  as follows: 
\begin{align*}
 V_{t}^{-1}&=\frac{1}{1-r}V_{t}^{-1}-\frac{r}{1-r}\frac{V_{t}^{-1}\bar{\vx}_t\bar{\vx}_t\T V_{t}^{-1}}{1-r+c_t},
\end{align*}
where $c_t=r\bar{\vx}_t\T V_{t}^{-1}\bar{\vx}_t$.

Finally the SDNML density function is written as follows:
\begin{align*}
 p_{\mbox{\tiny SDNML}}(x_t|x^{t-1})&=\frac{1}{K_t(x^{t-1})}\frac{s_t^{-(t-t_0)/2}}{s_{t-1}^{-(t-t_0-1)/2}},
\end{align*}
where the normalization factor $K_t(x^{t-1})$ is calculated as follows:
\begin{align*}
 K_t(x^{t-1})&=\frac{\sqrt{\pi}}{1-d_t}\sqrt{\frac{1-r}{r}}(1-r)^{-\frac{t-m}{2}}\frac{\Gamma((t-t_0-1)/2)}{\Gamma((t-t_0)/2)},
\end{align*}
with $d_t=c_t/(1-r+c_t)$.

\end{document}